\documentclass[runningheads]{llncs}
\usepackage{graphicx}
\usepackage{amsfonts,amsmath,amssymb} 
\usepackage{color}

\usepackage{booktabs}
\newcommand{\mbf}[1]{{\mathbf{#1}}}
\usepackage[usenames,dvipsnames,table]{xcolor}
\usepackage[width=122mm,left=12mm,paperwidth=146mm,height=193mm,top=12mm,paperheight=217mm]{geometry}
\usepackage{url}
\usepackage[breaklinks=true]{hyperref}
\usepackage{breakcites}
\usepackage{epstopdf} 
\usepackage{fancyhdr}

\begin{document}

\mainmatter
\title{~~\vskip0pt Extending Maps with Semantic and Contextual Object Information for Robot Navigation: a Learning-Based Framework using\\ Visual and Depth Cues}
\titlerunning{Semantic Map Augmentation for Robot Navigation}        

\author{Renato Martins$^{1,2}$ \and  Dhiego Bersan$^{1}$ \and Mario F. M. \\ Campos$^{1}$ \and Erickson R. Nascimento$^{1}$}

\authorrunning{Preprint version of \url{https://doi.org/10.1007/s10846-019-01136-5}} 

\institute{ 
	$^{1}$Universidade Federal de Minas Gerais (UFMG), Brazil \hspace{0.3cm}  $^{2}$INRIA, France
}

\maketitle
\pagestyle{headings}
\fancyhf{}
\lhead{{Preprint paper version to appear at Journal of Intelligent \& Robotic Systems, available online at: \url{https://doi.org/10.1007/s10846-019-01136-5}}}

\begin{abstract}
	This paper addresses the problem of building augmented metric representations of scenes with semantic information from RGB-D images. We propose a complete framework to create an enhanced map representation of the environment with object-level information to be used in several applications such as human-robot interaction, assistive robotics, visual navigation, or in manipulation tasks. Our formulation leverages a CNN-based object detector (Yolo) with a 3D model-based segmentation technique to perform instance semantic segmentation, and to localize, identify, and track different classes of objects in the scene. The tracking and positioning of semantic classes is done with a dictionary of Kalman filters in order to combine sensor measurements over time and then providing more accurate maps. The formulation is designed to identify and to disregard dynamic objects in order to obtain a medium-term invariant map representation. The proposed method was evaluated with collected and publicly available RGB-D data sequences acquired in different indoor scenes. Experimental results show the potential of the technique to produce augmented semantic maps containing several objects (notably doors). We also provide to the community a dataset composed of annotated object classes (doors, fire extinguishers, benches, water fountains) and their positioning, as well as the source code as ROS packages. \footnote{Preprint paper version to appear at Journal of Intelligent \& Robotic Systems, available online at: \url{https://doi.org/10.1007/s10846-019-01136-5}}
	
\end{abstract}

\section{INTRODUCTION}

Scene understanding is a crucial factor for the deployment of intelligent agents in real-world scenes in order to perform and support humans in everyday tasks \cite{safety19}. We have recently witnessed significant advances in the fields of scene understanding, human-robot interaction and mobile robotics, but they are still often challenged by typical adversities found in real environments. Surprisingly, these adversities are always (and successfully) handled daily by humans using mostly vision as primary sense. From their tender age humans learn to recognize and to build more abstract representations of what they observe in their surroundings: we look at with the eyes, but we see with the brain. 

In this context, embedding a higher level of scene understanding to identify particular objects of interest (including people), as well as to localize them, would greatly benefit intelligent agents to perform effective visual navigation, perception and manipulation tasks. Notably, this is a desired capability in human-robot interaction or in autonomous robot navigation tasks in daily-life scenes since it can provide ``situation awareness'' by distinguishing dynamic entities (e.g., humans, vehicles) from static ones (e.g., door, bench)~\cite{kit18,lane18}, or to recognize unsafe situations. This competence is also instrumental in the development of personal assistant robots, which need to deal with different objects of interest for guiding visually impaired people to cross a door, to find a bench, or a water fountain. Moreover, this higher level representation can provide awareness of dangerous situations (such as the presence ahead of steps, stairs) and of other people for safe navigation and interaction~\cite{pronobis12,papadakis17}. Recent advances of data-driven machine learning techniques and the increased computing capability of daily-use electronic devices have allowed envisaging transferring, to artificial agents, the human skills required to build these higher-level representations. It is then desirable to integrate these advances notably to mobile robotic systems, allowing them to perform more complex tasks, in safer conditions and in less specialized environments.

\begin{figure}[!t]
	\includegraphics[width=1\linewidth]{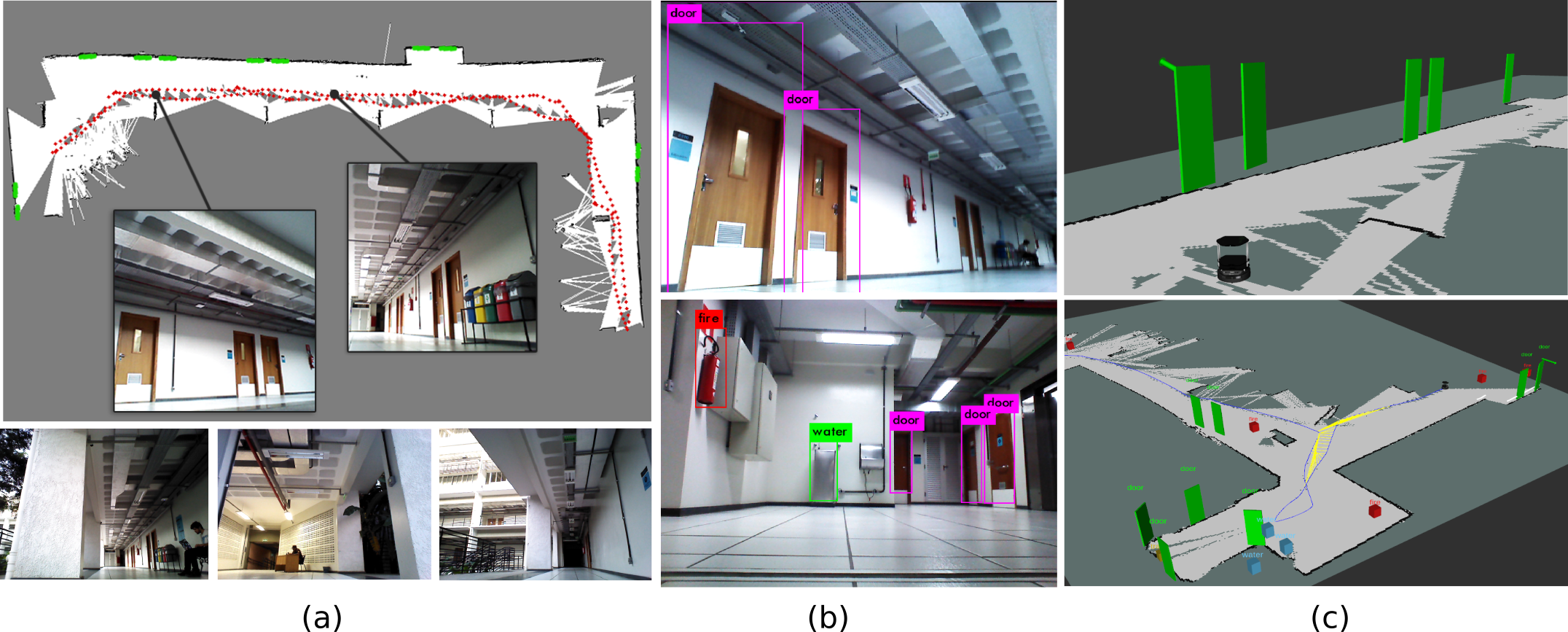}\vspace*{-0.2cm}
	\caption{{Augmented semantic mapping overview. (a) Bird's-eye view of the 2D map and door locations (in green) and some image frames of the dataset; (b) Object detection examples; and (c) Visualizations of the augmented semantic map output.}}\label{fig:intro}
\end{figure}

In this paper, we propose and evaluate a learning-based framework using visual and depth cues for building semantic augmented metric maps. The resulting representation combines both environment structure, appearance (metric map) and semantics (objects classes). The presented approach detects and generates models of objects in the surrounding environment using an RGB-D camera (or any stereo camera rig such as ZED 3D stereo cameras) as primary sensorial input. In a first moment, these RGB-D images are processed by a convolutional neural network to extract object classes as higher-level information, which is then leveraged by a localization and tracking system of the object instances over time. Finally, the environment representation is extended with the semantic information extracted using the object categories. In order to allow easy deployment in different robotic platforms, the full system is integrated in ROS (Robot Operating System).
Moreover, we also provide a dataset acquired in indoor environments with corridors and offices, containing annotated objects positions to help assessing and evaluating 3D object detection and mapping techniques. A characteristic result of our framework is depicted in Figure~\ref{fig:intro}, containing some image frames from one data sequence of the proposed dataset, as well as the respective object detections and augmented maps. 

\pagestyle{headings}
A preliminary conference version paper is introduced in our previous \cite{dbersan18}. In this manuscript, we have made a number of major modifications that we summarize as follows:

\begin{itemize}
	\item The localization and object tracking of the classes are improved to handle multiple objects per image and to support online pose updates, during loop closing of the localization and metric mapping back-end. 
	\item The object extraction and positioning are extended to support object classes with more complex shapes than planar patches. 
	\item Additional experiments and results are presented with object classes beyond ``door", such as ``person", ``bench", ``trash bin", among others.
	\item We present the training strategy and protocols used with the neural network for taking into account custom object classes. We also describe and provide the code for performing data augmentation and labeling, given a small sample of images from the additional object classes.
	\item We also provide the image training samples, source code, dataset sequences and video demos of the project \footnote{\url{https://www.verlab.dcc.ufmg.br/semantic-mapping-for-robotics/}}.
\end{itemize} 

The rest of this paper is structured as follows. In Section~\ref{sec:rel}, we discuss some recent related work on semantic object information and augmented map representations. Section~\ref{sec:met} presents the main stages of our semantic map augmentation. Then, we describe in Section~ \ref{sec:exp} the experimental setup, implementation details and the obtained results using real image sequences. The proposed dataset, that includes three data sequences collected in indoor scenes, is introduced in Section~\ref{sec:dataset}. Finally, Section~\ref{sec:conc} presents concluding remarks and discusses some perspectives of the work.

\section{RELATED WORK}\label{sec:rel}

There has been a great interest from the computer vision and robotics communities to exploit object-level information since from the perspective of many applications, it is beneficial to explore the awareness that object instances can provide for assistive computer vision \cite{perez14,leo2017computer,wang2019self}, tracking/SLAM \cite{li16semi,davidson17}, or place categorization/scene recognition and life-long mapping \cite{hane2016,wang18semantic}.

\subsection{Object Detection and Segmentation}
In order to build our extended map representation, we perform ``instance semantic" segmentation of objects leveraging an object detector with geometric priors.
Object-level representations are, in general, gathered from solving the challenging problems of object detection and semantic segmentation/labeling. An extensive amount of work have previously been reported to tackle these problems, employing a plethora of formulations ranging from graph-cuts, belief-propagation, or convex relaxation optimization/variational optimization, to name a few (the reader is referred to the survey \cite{seman2019survey}). However, the majority of recent state-of-the-art techniques are grounded on neural networks~\cite{alexnet12,rcnn15,shot15,yolo16}. 

Most recent object detection techniques are based on the generation of image region proposals, i.e., bounding boxes, and then predicting the most likely object class for each region. Commonly used benchmarks to evaluate object detection algorithms are the PASCAL Visual Object Classes (VOC) datasets~\cite{pascal10}, ImageNet~\cite{ILSVRC15} for object detection or the Multiple Object Tracking benchmarks (MOT) conceived specially to the evaluation of detection of humans in video. On the other hand, semantic segmentation is often done in the level of pixels and outputs different object classes in the image, but without object instance level information. Recent works as Mask-RCNN \cite{mask17} and YOLACT \cite{bolya-iccv2019} perform ``instance semantic" segmentation by combining several nets to simultaneously detect object instances and their semantic segmentation. The bottleneck of aforementioned approaches adopting supervised semantic segmentation is the user effort required to annotate pixel-wise a large number of images containing the classes of interest. Furthermore, these approaches have a high computational requirement, which limits the application to mobile robotic systems for real-time operation.  

Recent works in the area of intelligent vehicles \cite{gaidon2016virtual,cityscapes,zhan2018mix} presented databases of pixel-wise semantic segmented images with object classes such as pedestrians, road, sidewalk, car, sky. Also, some realistic image proxy engines have been proposed to overcome the annotation effort to segment some object classes in indoor scenes, as with the ScanNet dataset~\cite{dai2017scannet} or the Stanford 2D-3D-S dataset~\cite{armeni17}. However, concerning indoor visual navigation and assistive computer vision, the classes of interest such as doors, stairs or other path anomalies are not present or are not segmented in these datasets~\cite{wang2019self,perez14,salaris15}. Unfortunately, the majority of available datasets for both semantic segmentation and object detection do not explicitly consider these objects \cite{firman16}. In order to overcome this limitation, we acquired and labeled several images containing these custom objects of interest. In this paper, we adopted the object detection trend as a backbone for building our semantic-object augmented representation because the object information level met the expected augmented map requirements, but also because of the computational effort when making online inference with fully instance-segmentation networks. Furthermore, the required user annotation effort for pixel-wise labeling is also higher when compared to box object annotation.

\subsection{SLAM and Augmented Semantic Representations}

The combination of semantic information to support mapping and localization has been also explored by several recent works. For instance, Nascimento et al.~\cite{Nascimento:2011:IROSASP} applied a binary RGB-D descriptor to feed an Adaboost learning method to classify objects in a navigation task. McCormac et al.~\cite{davidson17} proposed a method for semantic 3D mapping. Their work combined the formulation of Whelan et al.~\cite{moreno15}, an RGB-D based SLAM system for building a dense point cloud of the scene, with an encoder-decoder convolutional network for pixel-wise semantic segmentation. The segmented labels are then projected/registered into the 3D reconstructed point cloud. Similarly, Li and Belaroussi~\cite{li16semi} provided a 3D semantic mapping system from monocular images. Their methodology is based on LSD-SLAM~\cite{engel14}, which estimates a semi-dense 3D reconstruction of the scene and performs camera localization from monocular images. Similarly to McCormac et al.~\cite{davidson17}, the metric map and the semantic labeling are combined in order to obtain the semantic 3D map. 

As discussed previously, due to runtime computational requirements and the amount of user effort to pixel-wise segmentation of the classes of interest for supervised semantic segmentation, we propose an instance semantic segmentation that leverages a lightweight data-driven object detection network with a model-based segmentation (object geometric shape priors). We also perform instance association and tracking through different time frames in order to build more complete and accurate extended maps.

\begin{figure}[!t]
	\centering\includegraphics[width=\linewidth]{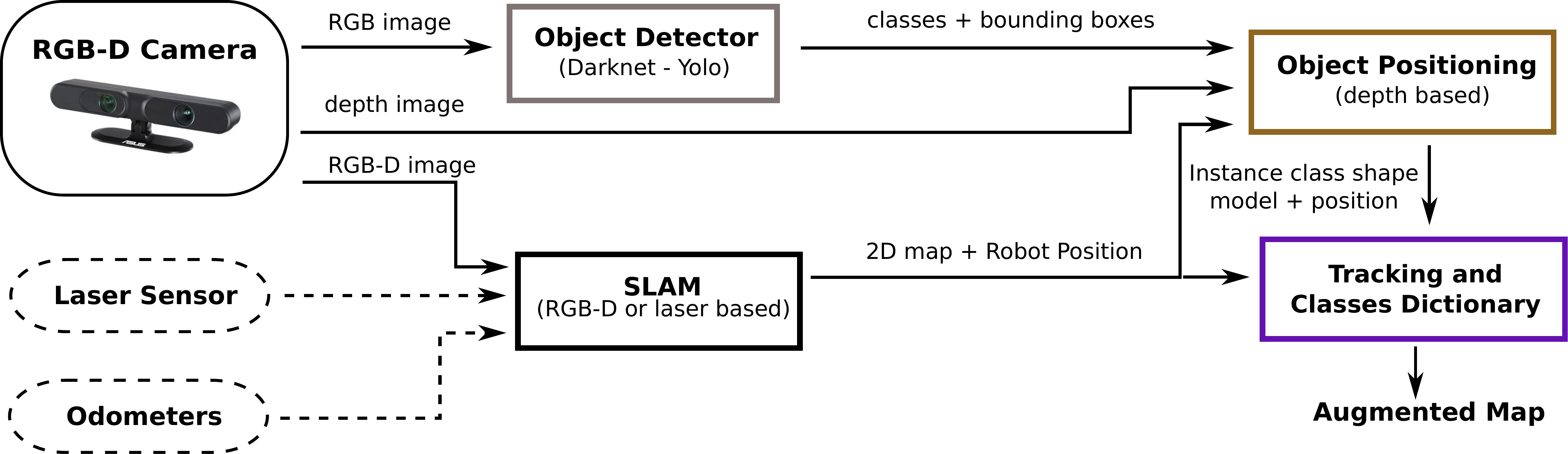}
	\caption{Visualization of the formulation pipeline, showing the main modules and some of the information exchanged between them.}
	\label{fig:p1}
\end{figure}

\section{METHODOLOGY}\label{sec:met} 

We divide our approach into four main components. An overview of the complete formulation is shown in Figure~\ref{fig:p1}. The first component addresses the semantic categorization and location of objects in the image, which employs a neural network to detect pre-trained object classes in real-time. This information is then used in a SLAM/localization step, which tracks the camera positioning in the scene and creates a projected 2D grid-based map of the environment using the available onboard robot sensors. Subsequently, we perform an efficient model-based object instance segmentation, from the object detection combined with 3D shape modeling priors. This component processes the information of the two previous components, together with the point cloud information, to localize the observed objects in the current frame and to segment pixels by fitting a primitive shape model (e.g., a planar patch for doors). Finally, the last component tracks previously localized objects in the map over time in order to combine multiple object measurements. 

\subsection{Visual Object Categorization and Detection}\label{sec:dete}

This section describes the first component of the augmented mapping framework. We start extracting a preliminary higher level representation of the scene with the detection of object classes of interest that are in the RGB-D camera field-of-view. For that, we profit of recent research progress on convolutional neural networks to reason from images to find objects and predict their semantic information, i.e., their location and category in the image. We selected the ``You only look once'' (Yolo) network~\cite{yolo16} among the various available object detection techniques \cite{alexnet12,rcnn15,resnet16}, because of its low computational effort and high precision-recall scores. The output of the network (as further described in the works \cite{dbersan18,yolo16}) are bounding boxes modeled with four parameters: the center position coordinates \((x,y)\), their width $w$ and height $h$.

\begin{figure}[!t]
	\centering
	\includegraphics[width=0.48\linewidth]{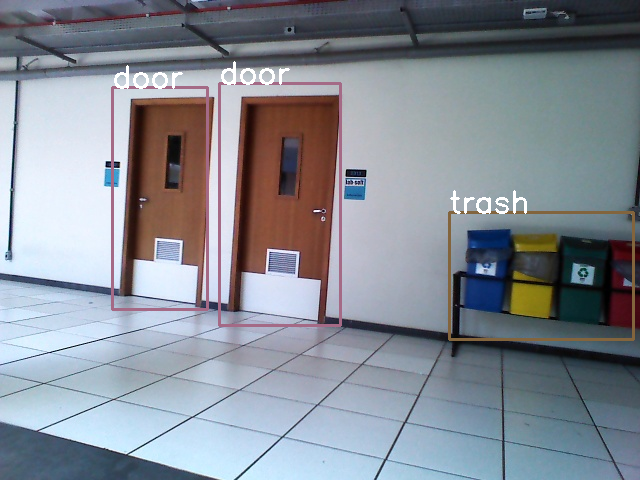}\hspace*{0.2cm}
	\includegraphics[width=0.48\linewidth]{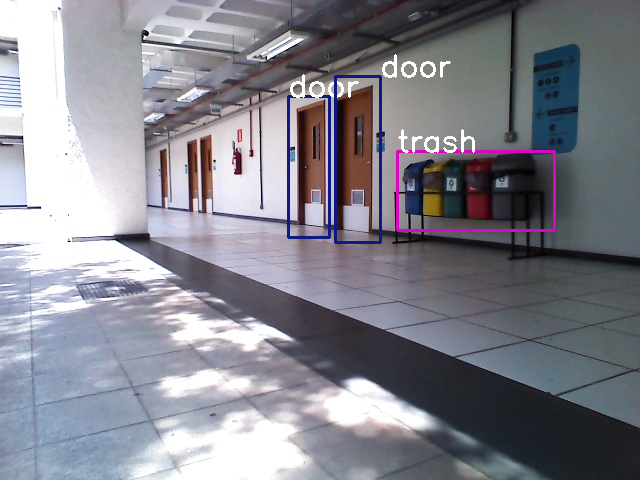}\\
	\includegraphics[width=0.48\linewidth]{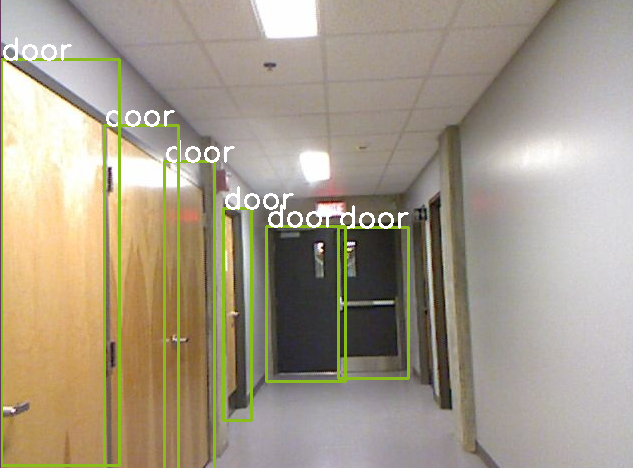}\hspace*{0.2cm}
	\includegraphics[width=0.48\linewidth]{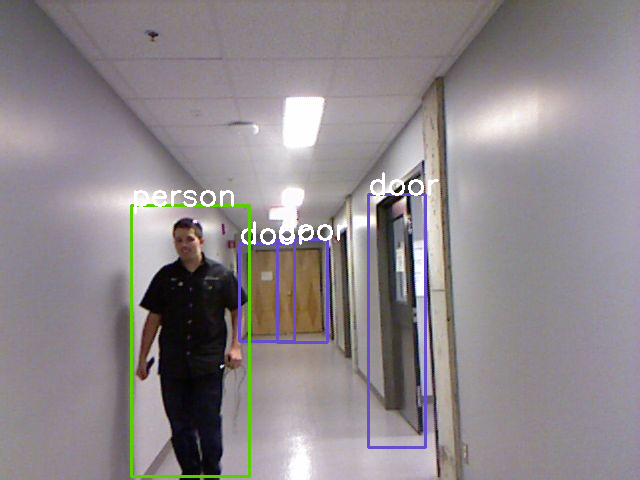}\\
	\caption{Visualization of the proposed labeling tool with different object classes. }
	\label{fig:label-tool}
\end{figure}

In our context of understanding and reasoning mainly in indoor scenes, the training images of Yolo contained classes such as ``door'', ``bench'', ``person'', ``water fountain'', ``trash bin'' and ``fire extinguisher'', as shown in Figure \ref{fig:label-tool}. Since the aforementioned available datasets \cite{firman16} did not contain annotated images with these classes (``door" images are available on ImageNet but still with high appearance variability), we need to label and perform data augmentation in order to successfully detect these objects.

\subsection*{Training and Dataset Augmentation}

We trained the network to detect a set of custom classes using a small amount of pictures taken from different objects in the environment, as well as using pre-labeled images from datasets available online (mainly for human detection). Our custom image labeling dataset consists of around $1,000$ pictures of \emph{doors, benches, trash bins, water fountains} and \emph{fire extinguishers}, together with the labels of their locations and classes. The labels were manually added using a tool developed for this purpose, which we also provide with the code. A preview of the labeling process is shown in Figure \ref{fig:label-tool}. Our tool is structured to make easier the annotation and network training.

For the detection of people, we adopted the Pascal VOC 2017 and 2012 datasets \cite{pascal10}. This dataset comprises about $20,000$ images of people, and their corresponding bounding boxes. The configuration files generated for each subset (from our custom object images and the Pascal VOC dataset) were combined in order to train the network considering objects from both datasets. The network architecture was redefined to have the number of classes updated, as well as the number of filters.

\begin{figure}[!t]
	\centering
	\includegraphics[width=0.32\linewidth]{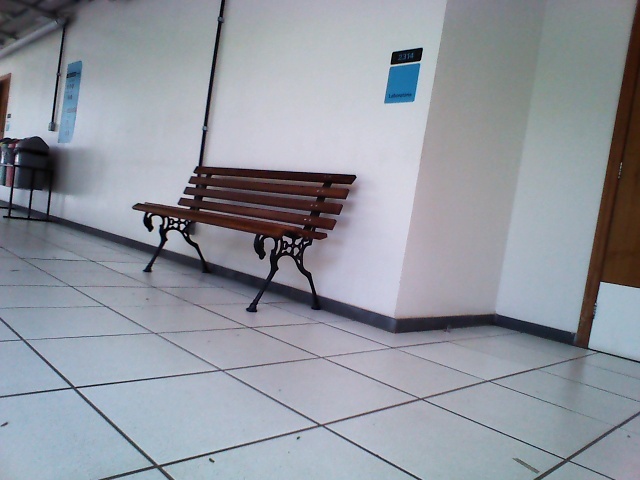}\hspace*{0.15cm}
	\includegraphics[width=0.32\linewidth]{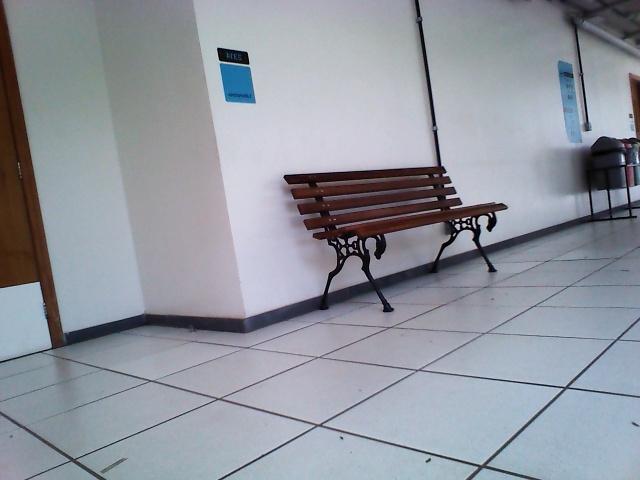} \hspace*{0.05cm}
	\includegraphics[width=0.32\linewidth]{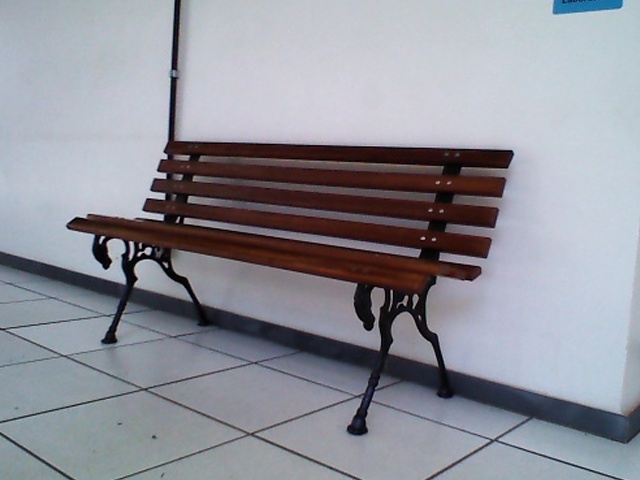}
	\caption{Dataset augmentation operations. The first image corresponds to the original frame, the resulting flipped image (in the center) and scaled \& translated image (at right). }
	\label{fig:augmentation}
\end{figure}

\begin{figure}[!t]
	\includegraphics[width=\linewidth]{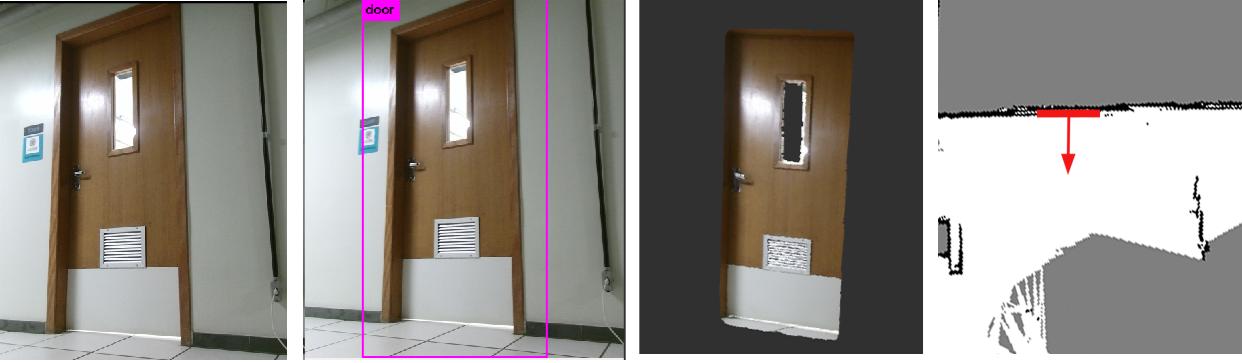}
	\caption{Door detection: input image, object detection bounding box, RANSAC inliers for planar segmentation, object model represented in the map.}
	\label{fig:obj-detection}
\end{figure}

One issue encountered after training and testing the network was that Pascal VOC datasets have an overwhelmingly more significant number of images than our custom built dataset of other objects. This led the network to become biased towards the \emph{person} class, while rarely detecting other objects. To overcome this issue, we augmented the custom-built dataset using common dataset augmentation operations such as flipping, scaling \& translating, and adding intensity noise. Each operation doubled the number of images. We then applied two random noise levels, two scale, and one flipping operations to the original images, which increased the number of training images by a factor of $2^5-1=31$. An example of the augmentation result can be seen in Figure \ref{fig:augmentation}. The noise operation, although increasing the network robustness, did not significantly affect the image appearance to the human eye. The final output of the system is the object boxes (encoded by five parameters) at an average mean frame-rate of 30Hz with a Nvidia GeForce GTX 1060. Some detection prediction examples can be seen in Figures~\ref{fig:intro} and \ref{fig:obj-detection}.

\subsection{Localization and Mapping}\label{sec:loc}

Concurrently to the object detection described in Section~\ref{sec:dete}, we generated an initial 2D projected map representation of the environment, along with the localization of the robot in this representation. A plethora of techniques can be used to localize the robot, depending mostly on the available sensors and computational requirements. We set as the minimal required sensor setup to our system as one RGB-D camera, which information is exploited in all stages of the formulation. However, it is worth noting that the proposed localization module is also designed to consider LIDAR and wheel odometers sensors when these are available in the robotic system. Thus, three main setups are supported: 
\begin{itemize}
	\item[i)] The laser scan is not available. In this case, the depth image is sampled from the RGB-D camera in order to create the scan stream.
	\item[ii)] Odometers are not available.  The registration between the RGB-D frames is used in order to build the odometry information.
	\item[iii)] Both LIDAR and odometers are not available. We follow as indicated in the two previous i) and ii) settings.
\end{itemize} 
\begin{figure}[!t]
	\centering
	\includegraphics[height=4cm]{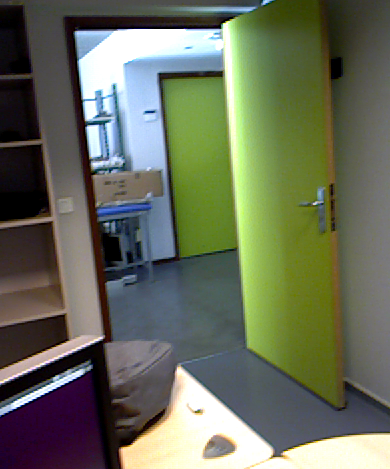}
	\includegraphics[height=4cm]{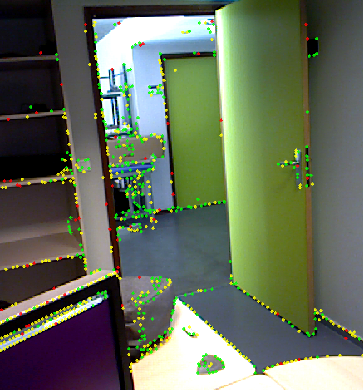}
	\includegraphics[height=4cm]{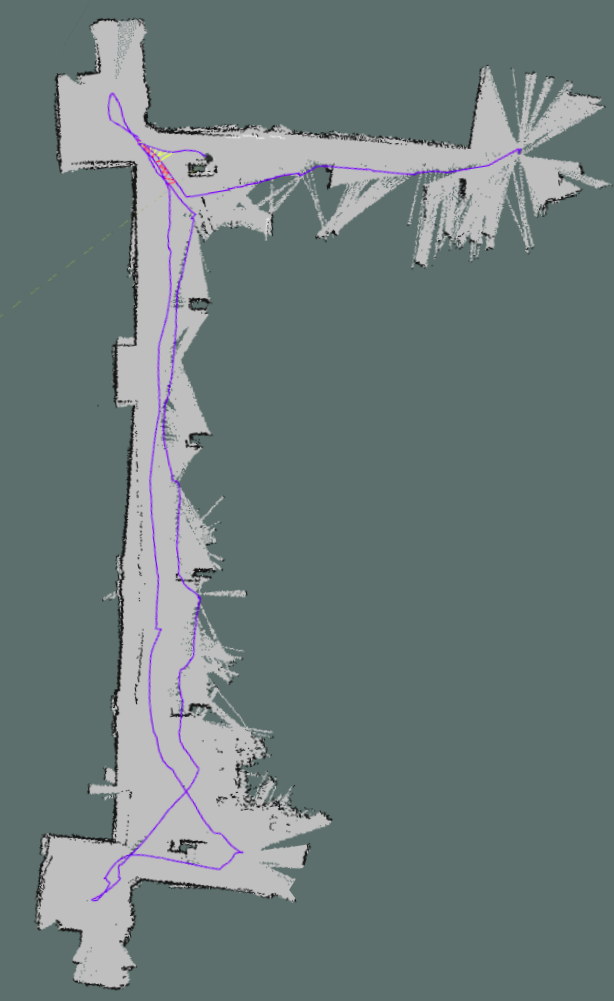}
	\includegraphics[height=4cm]{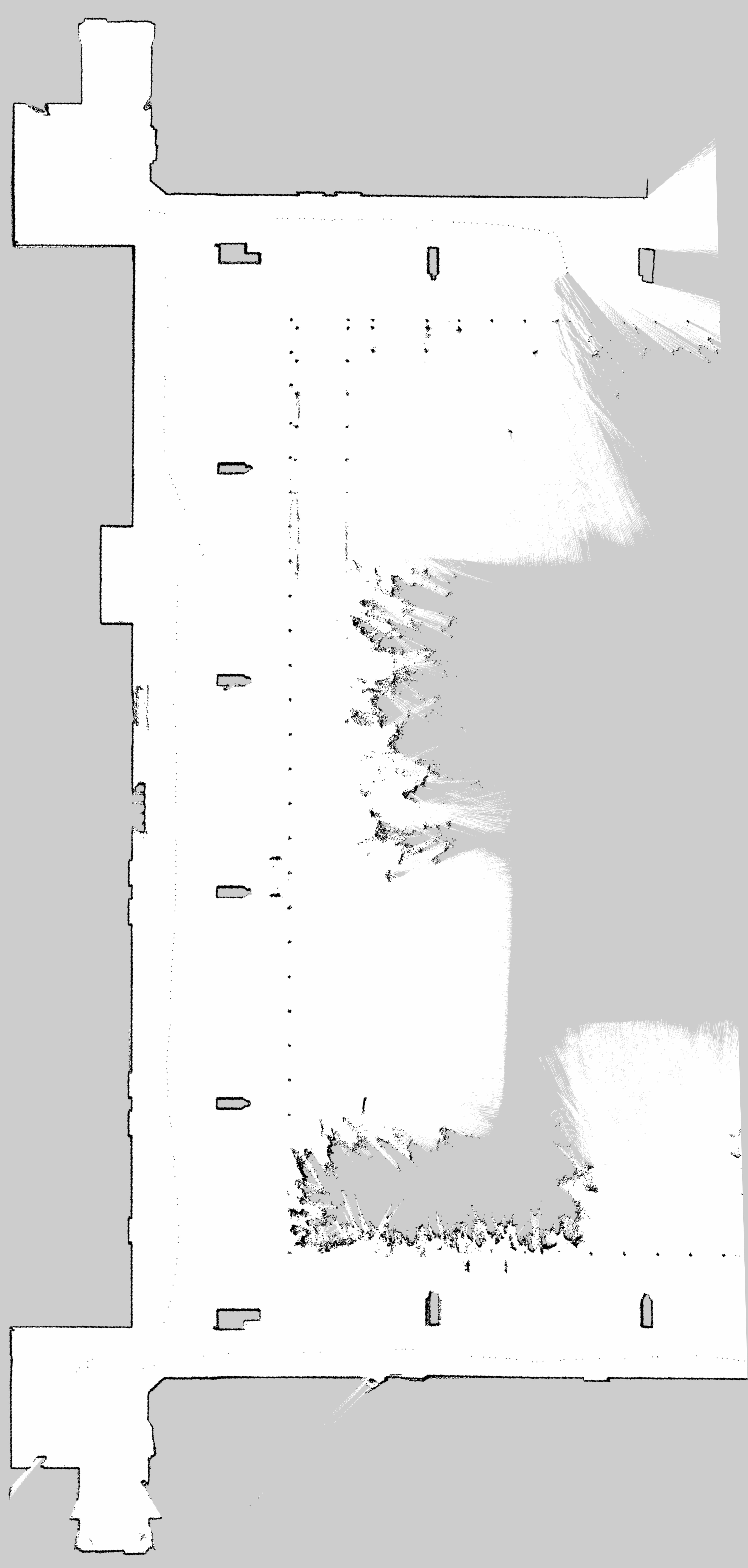}\\
	\includegraphics[width=0.96\linewidth]{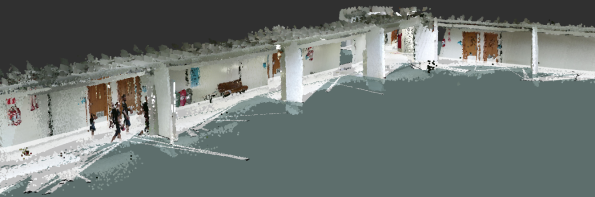}
	\caption{2D grid-based and textured point cloud of our first dataset sequence. Top row (from left to right): a sample input image, the extracted features from RTAB-Map, the resulting metric map and the map ground truth. Bottom row: the final textured 3D point cloud on top of the 2D grid-based representation.}
	\label{fig:localization_orb}
\end{figure}
In order to have an easy deployment system, we considered mostly SLAM/mapping techniques currently available on ROS. Also with the purposes of flexibility and portability, the adopted localization/mapping backbone is selected to produce an output consisting of a 2D grid-based representation of the environment $\mbf M$ along with the 2D projected location $\mbf{x}_r\in \mathbb{R}^3$ of the robot in the map: 
\begin{equation}
	\mbf{x}_r=(x,~y,~\theta)^T,
\end{equation} 
\noindent where $(x,y)$ is the position and \(\theta \) the orientation. This step can exploit any range-based or visual-based localization/SLAM algorithm, notably the provided framework supports and was tested with techniques already available on ROS as Gmapping SLAM \cite{gmapping07}, AMCL \cite{fox1999monte} and RTAB-Map library \cite{rtab19} which was initially developed for appearance-based loop closing and memory handling for large-scale scene mapping. These libraries provide localization and mapping techniques for several sensory modalities, including RGB-D, stereo or monocular camera settings for both 2D grid-based representation and the 3D textured point cloud of the scene. 

We remark that other state-of-the-art image registration techniques such as the feature-based ORB-SLAM \cite{orbslam15} and appearance-based RGBDSLAM~\cite{endres14} could also be used with minimal effort in the system, as long as the system provides camera localization and the 2D projected grip map of the scene. The required changes are then mainly in adjusting the API and ROS message exchange (subscribing and publishing topics) as done for AMCL, Gmapping and RTAB-Map algorithms. After performing several experiments, we adopted RTAB-Map for giving the most accurate and complete map results, as shown in the metric maps generated from the provided dataset sequences in Figure \ref{fig:localization_orb}. Finally, it is worth noting that one could also leverage the redundancy given by the available sensor settings, especially concerning laser information with the depth provided by the RGB-D camera. Also, the odometry information can be either gathered from encoders, range or visual information, which of these having their complementary properties, advantages and cons to the localization and mapping. 

\subsection{Model Fitting and Positioning}
Given a set of detected objects, we perform efficient object instance segmentation of nearly thin or flat objects by adopting primitive 3D shape priors. For instance, a plane is a reasonable primitive for representing ``doors''. From the RGB-D camera calibration parameters, we then reconstruct and find all 3D points inside the detected box, where the primitive model of the objects is fitted. The clustering technique adopted in the shape model fitting, to all classes except doors, was the Euclidean region growing segmentation technique \cite{Rusu_ICRA2011_PCL}, returning the centroid and respective convex hull dimensions. Whenever the detected objects are labeled as ``door", we fitted a planar patch using RANSAC \cite{ransaccomp} for estimating the position and orientation.  

\begin{figure}[!t]
	\includegraphics[height=0.17\linewidth]{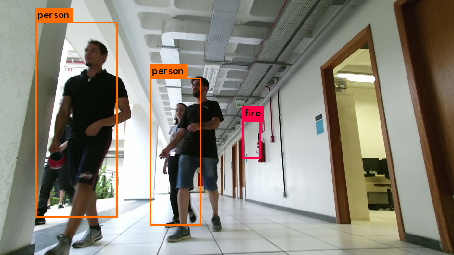}
	\includegraphics[height=0.17\linewidth]{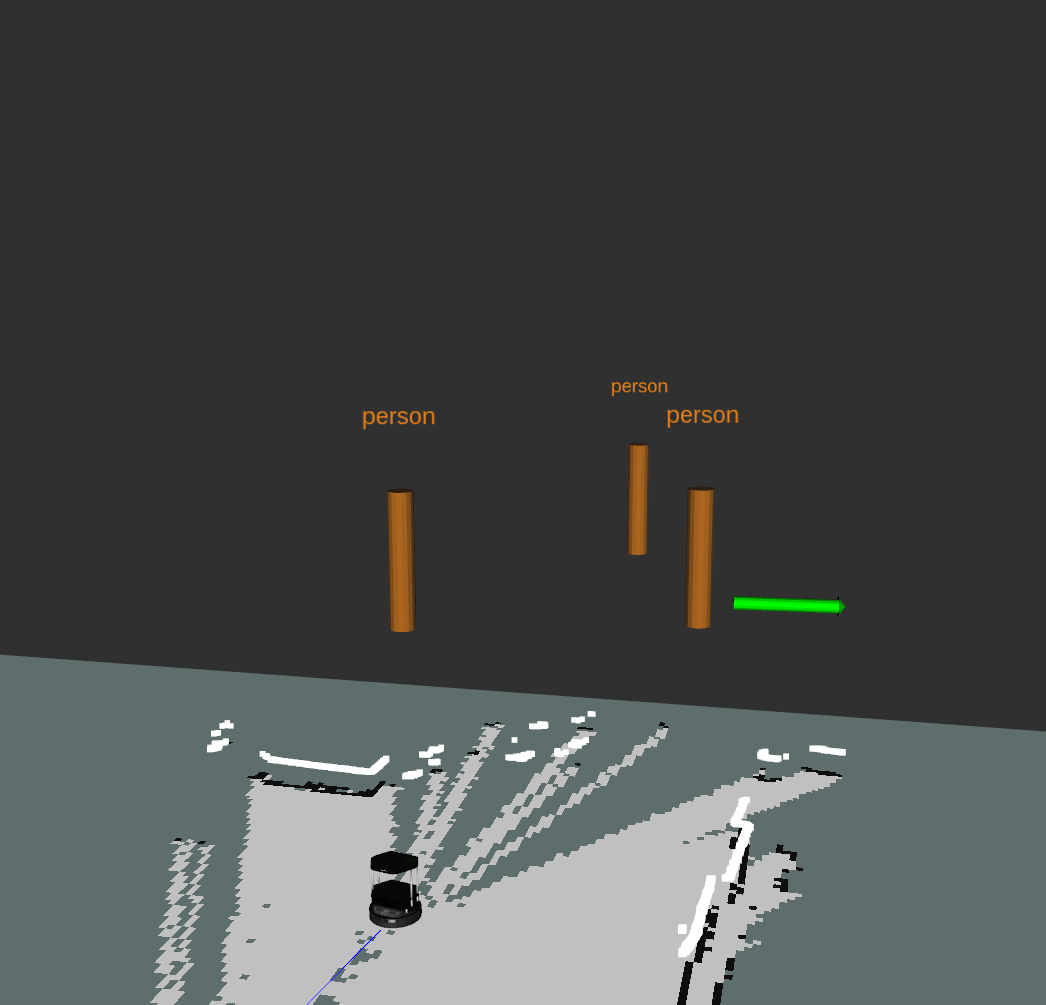}
	\includegraphics[height=0.17\linewidth]{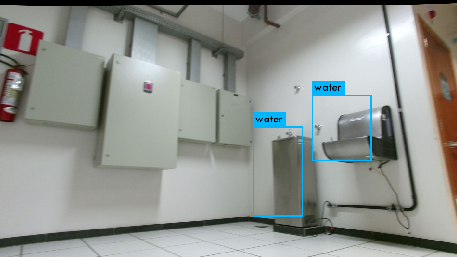}
	\includegraphics[height=0.17\linewidth]{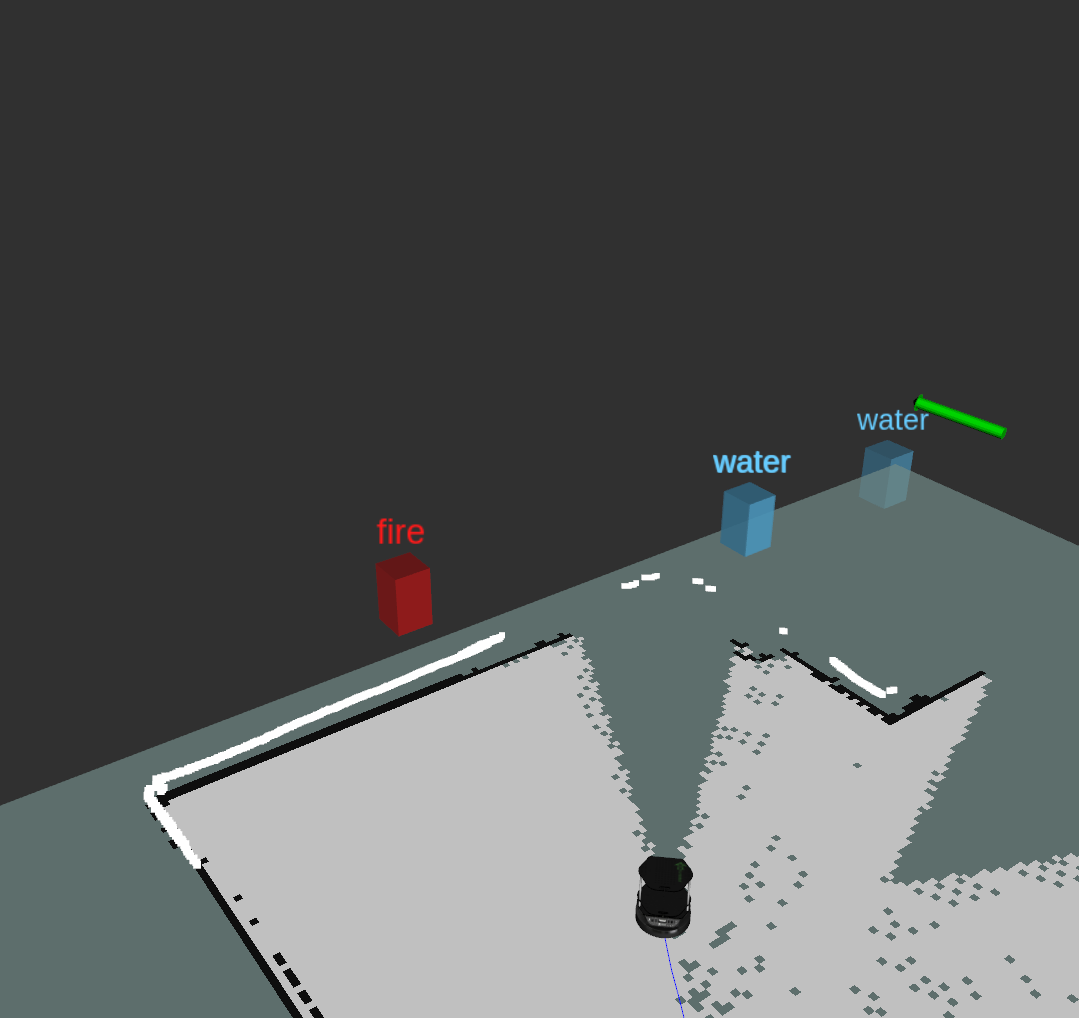}
	\caption{``Person", ``water fountain" and ``fire extinguisher" object detection and model fitting: input images and object model represented in the map.}
	\label{fig:obj-detection-new}
\end{figure}

The projected pose of each object, denoted by $\mbf y \in \mathbb{R}^3$, is then represented by the 2D projected centroid from the camera coordinate system to the global map coordinate system and its orientation. These steps can be seen in Figures~\ref{fig:obj-detection} and \ref{fig:obj-detection-new} for different objects.

\subsection{Object Tracking and Final Augmented Representation}

After observing and projecting objects onto a location on the map, the final step is to perform tracking of the captured objects. That is, given multiple observations of multiple classes of objects (doors, benches, trash bins, etc..) across different instants of time, we wish to infer which objects have already been observed before and which have not. Ideally, we want to associate every previously seen instances with the right stored instance, and unseen objects as new instances. This would allow us to augment the map with the correct information about the semantics of the environment. Erroneous associations on this step yield undesired results as multiples instances of the same object (false positives) or associating two different observations of two different objects as belonging to the same object (false negatives). 

Although tracking people is essential in re-active navigation and situation awareness, in this work, we do not perform the dynamic tracking of people because we are mostly interested to the static objects in the final augmented map representation. Furthermore, it is worth noting that an association/correspondence strategy based only on object locations is likely to fail to track humans. In this case, more elaborated models considering explicitly the appearance should be taken into account as, for instance, using bi-directional long short-term memories to handle appearance changes \cite{sadeghian2017tracking,kim2018multi}.

For any given frame, all the $m$ observed objects' positions of a given class are stored as a set of observations $\mbf Y = \{\mbf y_0,\mbf y_1, ..., \mbf y_m\}$. We want to compare and check if any of these observations match one in the dictionary of $n$ already observed instances of that same object class, $\mbf X = \{\mbf x_0,\mbf x_1, ..., \mbf x_n\}$. The association cost matrix $\mbf{D}({\mbf x_i, \mbf y_j})$ between both sets is computed using the Mahalanobis distance for every possible match: 
\begin{equation}
	\mbf{D}({\mbf x_i, \mbf y_j}) = \sqrt{\mbf(\mbf y_j - \mbf{x}_i)^T\mbf S_i^{-1} (\mbf y_j - \mbf{x}_i)},
\end{equation}

\noindent where $\mbf{x}_i$ is the i-th model of the $n$ matched instances ($i = \{1,2,3,...,n\}$) of  $\mbf X$ and $\mbf S_i$ is its related covariance matrix. Once the cost matrix is computed, the association between the observations and the dictionary instances are gathered from the Hungarian-Algorithm \cite{kuhn1955hungarian}. All the resulted associations which distances are smaller than a threshold $(\mbf{D}({\mbf x_i, \mbf y_j})<\delta)$ are assumed to correspond to previously seen objects; otherwise, new object instances representing the remaining observations are included in the dictionary. 

In order to track and to increase the accuracy of detected instances, each stored semantic object is modeled with a constant state Kalman filter \cite{dbersan18}, since we are interested in storing mostly static classes in the final augmented map, to maintain its state up-to-date and combine different objects observations.
\begin{figure*}[t!]
	\includegraphics[height=3cm]{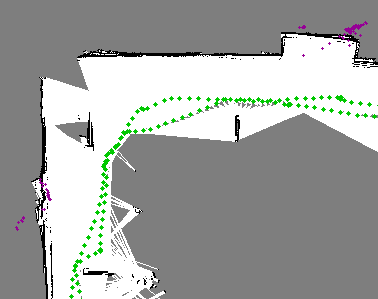}\hspace*{0.3cm}
	\includegraphics[height=3cm]{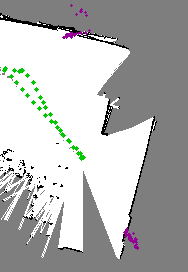} \hspace*{0.3cm}
	\includegraphics[height=3cm]{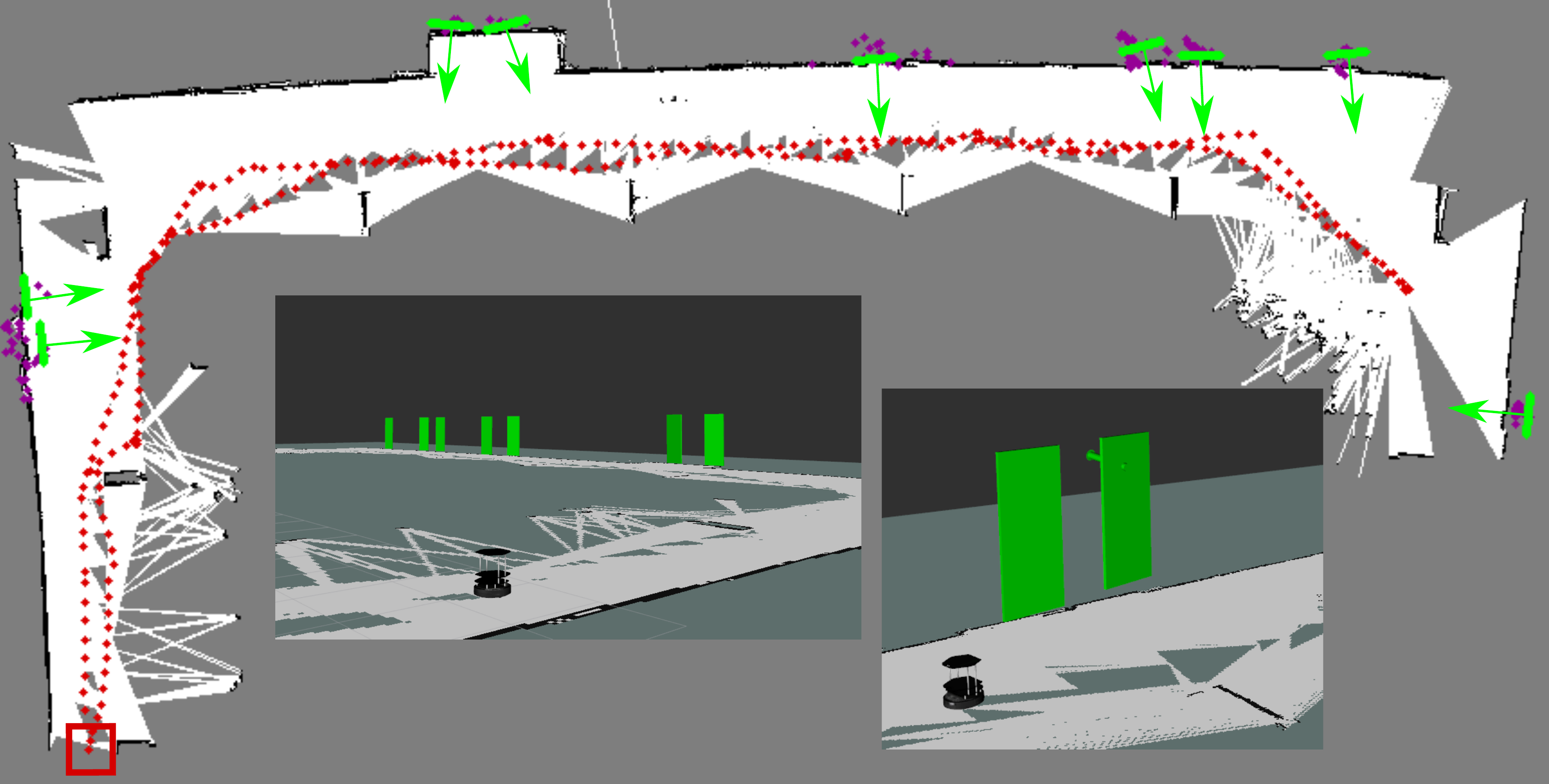}
	\caption{Door object observations during the robot navigation. The extracted positions \textbf{y} of the doors observed over time are shown in pink and the final tracked/filtered instances \textbf{x} are indicated in green. The left image also includes two visualizations of the augmented semantic map.}
	\label{fig:kalman}
\end{figure*}
Each filter combines the information of the different observations temporally as shown in Figure~\ref{fig:kalman}. The filter initialization and tunning details are described in Section~\ref{sec:param}. 

The advantage of this simple tracking approach is that it pays the way for the integration of different object models that can be sufficiently described from a positioning/geometric point of view in the scene. Specifically, the positional properties of the object models of interest to this work were sufficiently discriminant to perform the tracking, as long as the accuracy of the localization/mapping system, described in Section \ref{sec:loc}, was bellow the Mahalanobis distance threshold. 

\section{EXPERIMENTS}\label{sec:exp}
The experiments were performed online with a mobile robot navigating indoor scenes and offline using  previously acquired indoor dataset sequences. We also present qualitative results with a publicly available RGB-D dataset. We first detail the parameters setup considered in the experiments, and then we present some extended mapping results.

\subsection{Dataset and Object Training Samples}\label{sec:dataset}

\begin{figure}[t]
	\includegraphics[width=.9\linewidth]{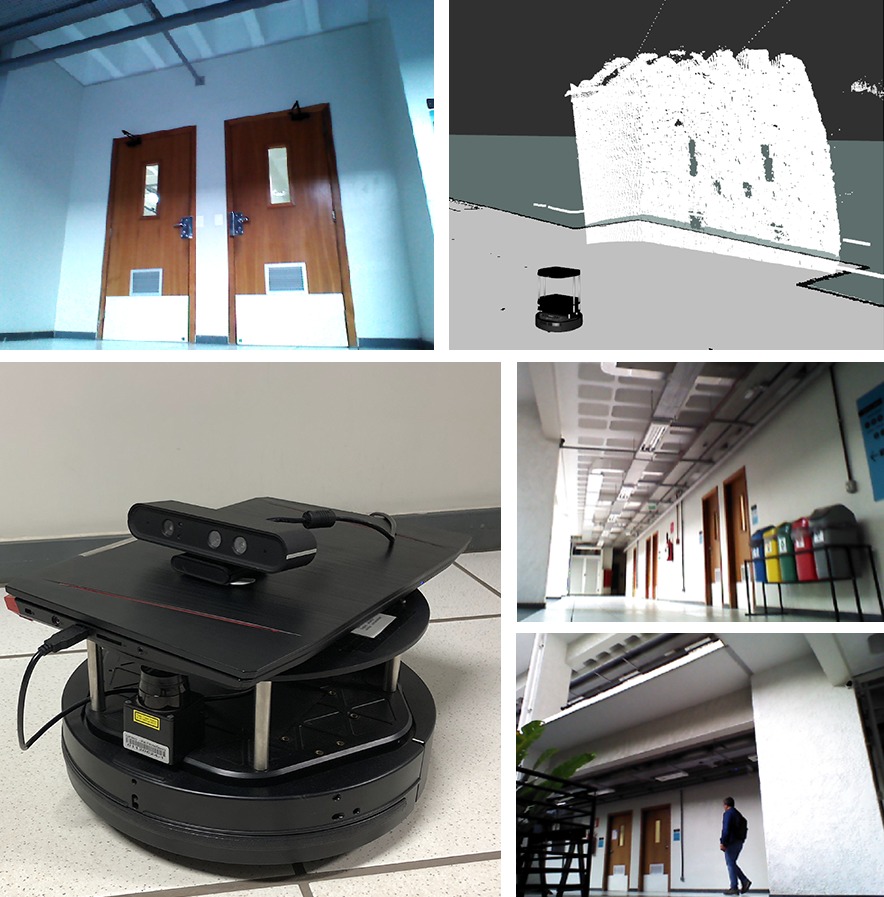}
	\centering\caption{Examples of scenes contained in the first sequence of the dataset and the Kobuki base robot. The first row displays and RGB frame and its corresponding point cloud visualization. The robot with on-board sensors (RGB-D camera and 2D LIDAR) is shown in the bottom left image. }
	\label{fig:robot}
\end{figure}

\begin{figure*}[!t]
	\centering\includegraphics[width=1\linewidth]{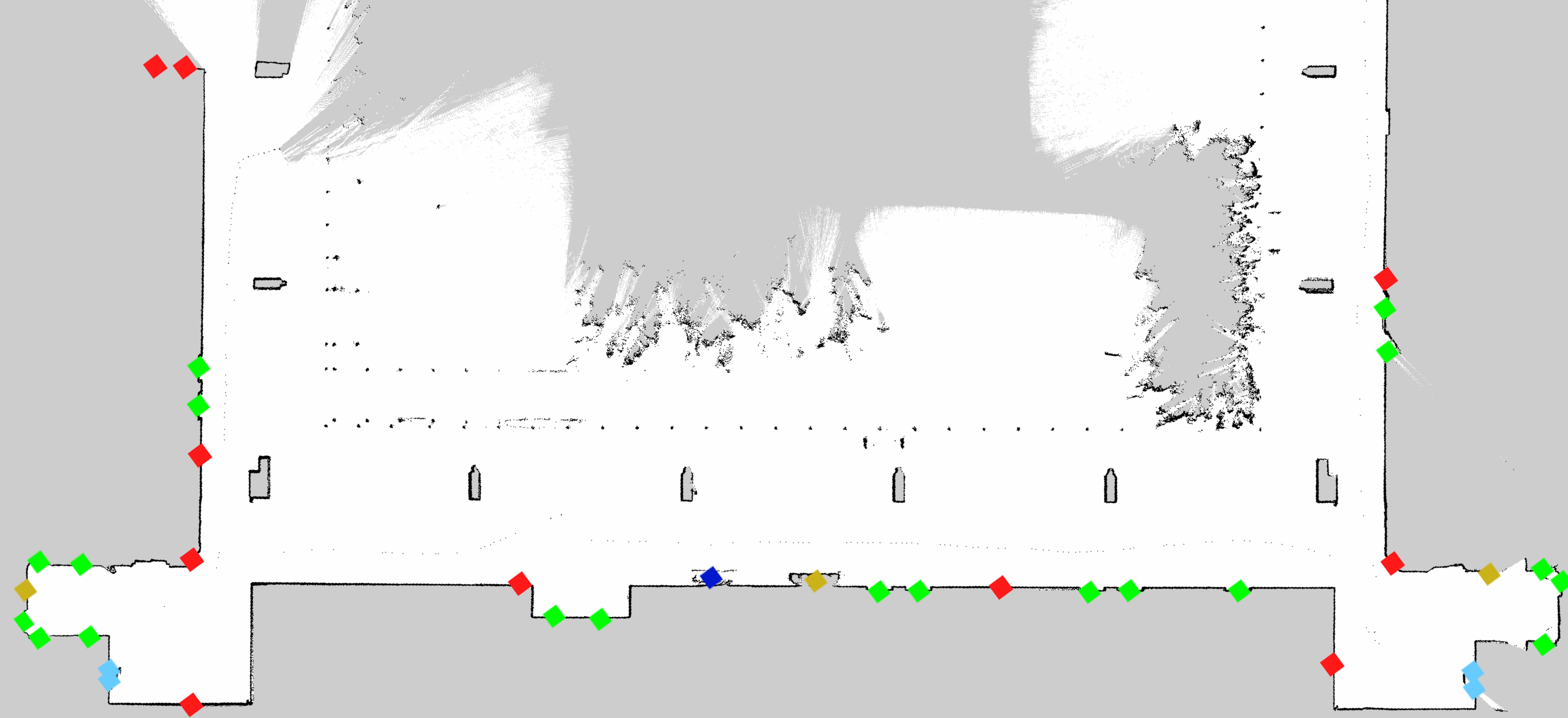}
	\caption{2D ground-truth map with the projected object positions used in the dataset sequences \textbf{sequence1-Kinect} and \textbf{sequence2-Astra}, and with a mapped area of $42m \times 18.5m$: doors (green squares), fire extinguisher (red squares), trash bin (in yellow, water fountain (in light blue) and bench (in dark blue).}
	\label{fig:ground-truth-map}
\end{figure*}

\begin{figure*}[!t]
	\centering\includegraphics[width=1\linewidth]{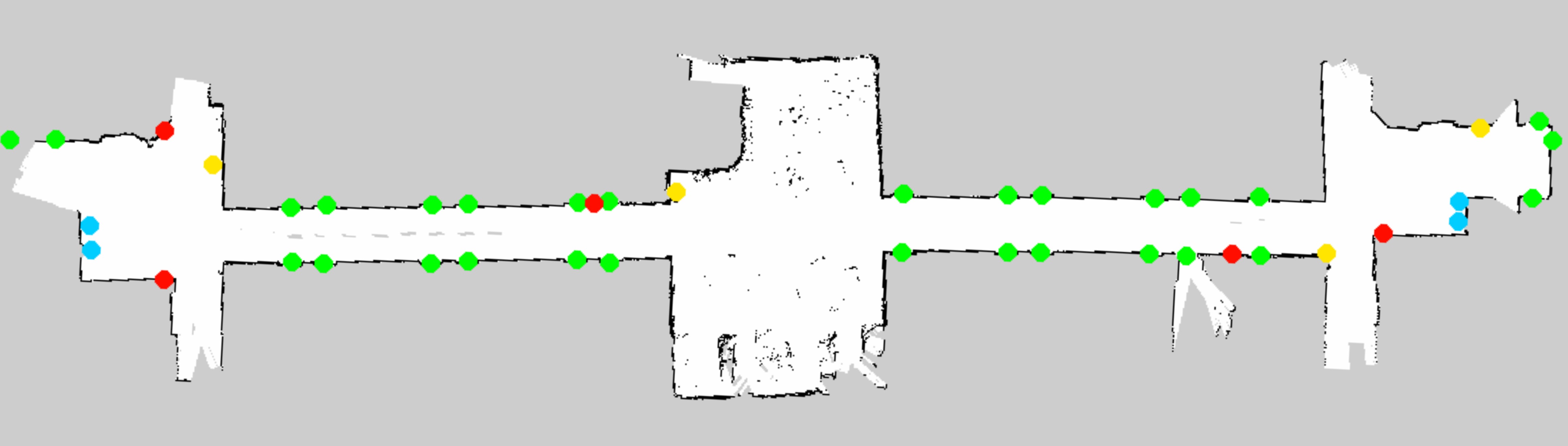}
	\caption{2D ground-truth map (covered mapped area of $54m \times 12m$) with the projected object positions used in the dataset sequence \textbf{sequence3-Astra}: doors (green squares), fire extinguisher (red squares), trash bin (in yellow, water fountain (in light blue) and bench (in dark blue).}
	\label{fig:ground-truth-map-7th}
\end{figure*}

Apart from online experiments and to evaluate the performances of the proposed method in controlled conditions, we collected a dataset containing three data sequences of different indoor places. These sequences were acquired while a robot was teleoperated in indoor environments, as depicted in Figure \ref{fig:robot}. Each sequence contains raw sensor streams recorded using the rosbag toolkit from two different RGB-D cameras, LiDAR and odometry. All data sequences contain different classes of objects: \textit{person, door, bench, water fountain, trash bin, fire extinguisher}, as shown in the images of Figures \ref{fig:intro}, \ref{fig:label-tool} and \ref{fig:robot}. Every class considered static (i.e., all, except for person and chair) have their location specified in a ground truth map we provide, as shown in Figure \ref{fig:ground-truth-map}. An overview of these three sequences is depicted in Figures \ref{fig:ground-truth-map} and \ref{fig:ground-truth-map-7th}, which also contain the projected object positions. The RGB-D sequences used two different RGB-D sensing cameras: Microsoft Kinect (\textbf{sequence1-Kinect}) and Orbbec Astra (\textbf{sequence2-Astra} and \textbf{sequence3-Astra}). Further details of time duration, data statistics and information parsers is given in the project page\footnote{\url{https://www.verlab.dcc.ufmg.br/semantic-mapping-for-robotics/}}.

As previously mentioned, this dataset was built since the majority of available datasets for both semantic segmentation and 3D object detection did not consider doors and the other objects of interest to the navigation in our indoor scenes. Unfortunately, a motion capture system was not available to get the precise camera position along with the displacement in all the covered area of the scenes, nor a fine-detailed 3D mesh reconstruction of the environments due to their extension. To circumvent this limitation, we obtained the 3D robot position and of objects for each sequence performing a fine-level localization on the 2D CAD model of the scene. We then computed the 3D position of each object relative to the local image frames.


\subsection{System Setup and Implementation Aspects}\label{sec:param}

We used a robotic platform containing a Kobuki base, where the different RGB-D cameras and LIDAR sensors were mounted, as described in the dataset Section~\ref{sec:dataset}. All the components of the formulation are integrated with ROS (Robot Operational System) and the output map generation runs at $15$ Hz in a laptop with Ubuntu 16.04, Intel core i7 and Nvidia GeForce 1050 Ti. Since our main goal is to extend maps with relevant object information that do not usually change position over time, interesting candidates for navigation and user interaction available in your sequences were doors, bench, water fountain, and fire extinguisher. To this end, we trained the network following the protocol indicated in Section \ref{sec:dete}. In the robot localization and mapping, we adjusted few parameters from the RTAB-Map default parameters (which are beyond 100), such as Reg/Strategy to use visual and depth information in the localization.

The geometric model fitting was performed with RANSAC \cite{Rusu_ICRA2011_PCL}. We allow the point to plane fitting to optimize coefficients, and the distance threshold to \(0.03\). 
From our experiments, this value accounted for errors in the camera depth images, while allowing a correct segmentation of door points from the wall, in case these lie in different planes. 
In the object association and tracking, we adopted a constant uncorrelated noise affecting the process and observation measurements (i.e., the error covariance are diagonal matrices). 

\subsection{Augmented Mapping Results}

\begin{figure*}[!t]
	\centering\includegraphics[width=.8\linewidth]{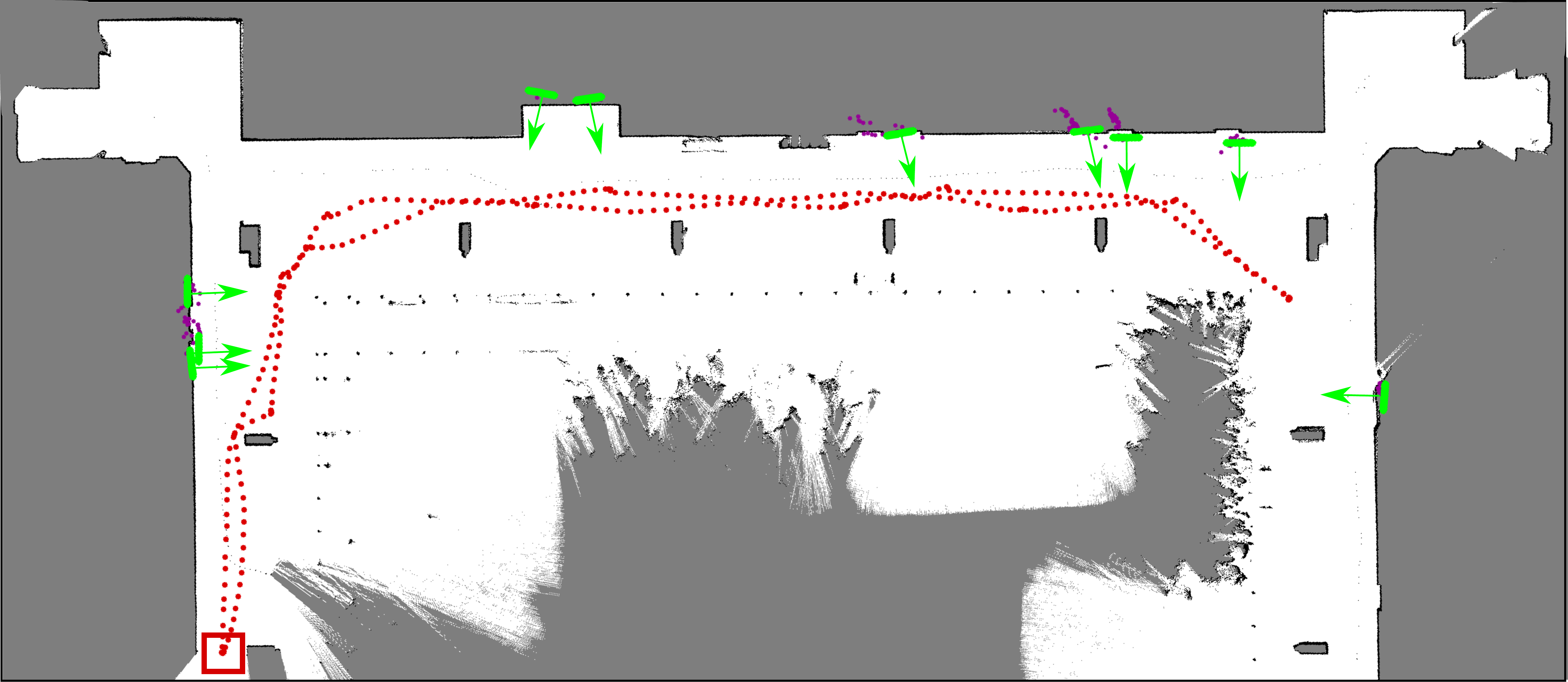}
	\caption{Augmented 2D map with door instances with localization-only (figure axes dimensions $55.7 \times 24.3m$). The red square indicates the starting and ending point of the robot trajectory (red-dotted). Purple dots are the unfiltered positions observations and green lines are the doors filtered results. The reconstructed map depicts with green arrows the position and orientation of objects.}
	\label{fig:results1}
\end{figure*}
\begin{table}[t!]
	\centering
	\caption{Object detection and tracking results of sequence \textbf{sequence1-Kinect}.}
	\begin{tabular}{lrrrr}
		\toprule
		\textbf{class} & \textbf{detection} & \textbf{FP} & \textbf{FN} & \textbf{avg. error {[}m{]}}\\ 
		\toprule
		door &	19 &	1 &	3 &	0.78\\
		bench &	1  &	1 &	0 &	1.2\\
		trash bin &	3 &	1 &	0 &	1.04 \\
		fire exting. &	9 &	1 &	3 &	0.53 \\
		water fount. &	4 &	0 &	0 &	0.61 \\
		\bottomrule
	\end{tabular}
	\label{table-no-kine}
\end{table}

\begin{figure*}[!t]
	\centering  \includegraphics[height=3.76cm]{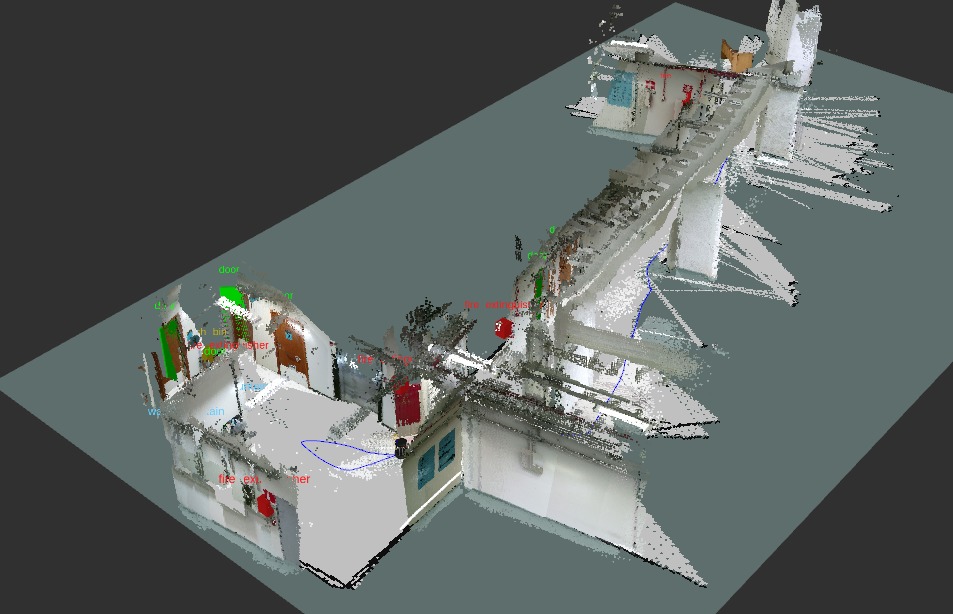} \includegraphics[height=3.76cm]{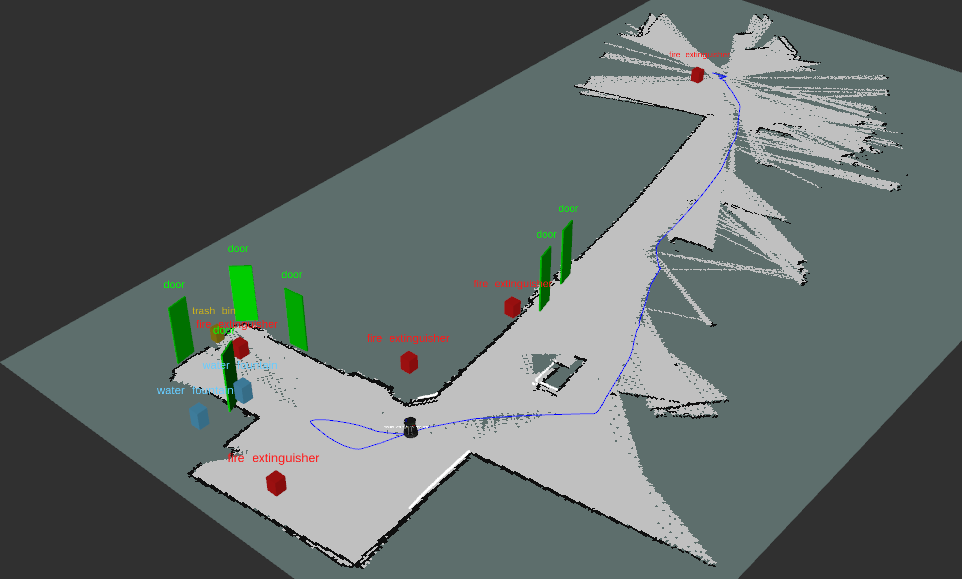} \\
	\includegraphics[width=1\linewidth]{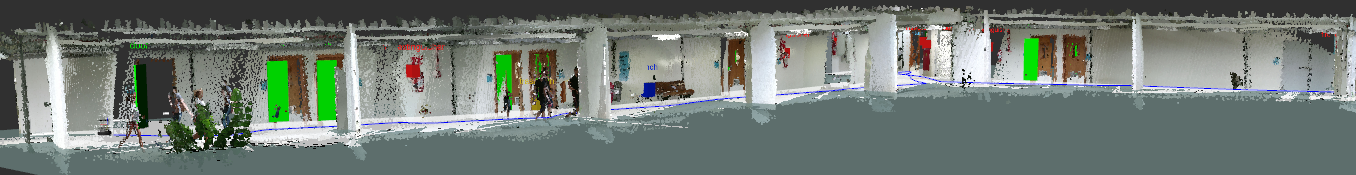}\\
	\includegraphics[width=1\linewidth]{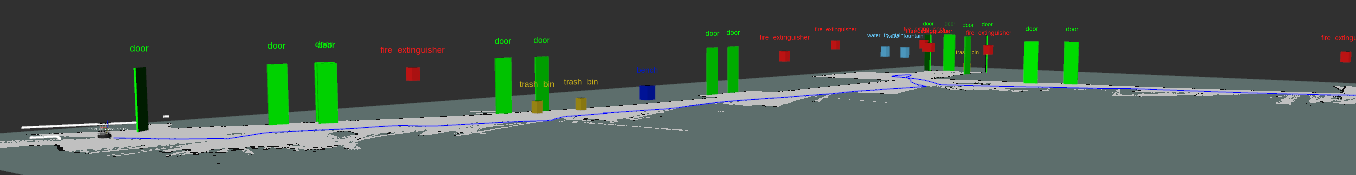} 
	\caption{Visualizations of the augmented map from sequence \textbf{sequence1-Kinect} with RTAB-Map. The geometric object primitives are shown in green, red and blue representing the ``door", ``water fountain" and ``fire extinguisher" respectively.}
	\label{fig:res-rtab1}
\end{figure*}

To exemplify the flexibility of the approach, the evaluation is done with two different localization strategies: one performing 3D RGB-D SLAM (RTAB-Map) and one 2D probabilistic re-localization approach in a previously generated map (Adaptive Monte Carlo localization - AMCL\cite{fox1999monte}). We show some of the extended map results for both of these approaches in Figures \ref{fig:results1}, \ref{fig:res-rtab1}, \ref{fig:res-rtab2} and \ref{fig:res-rtab3}. The detected objects are shown in green, red and blue representing ``door", ``water fountain" and ``fire extinguisher" respectively. The quantitative metrics adopted are the amount of false positives, which indicates percentage of objects that were wrongly instantiated, and the amount of false negatives that indicates the number of objects that were not integrated in the final map representation. The position errors of the objects is also considered. The adopted qualitative metric is the visual quality of the augmented semantic visualizations of the different scenes.

\begin{table}[t!]
	\centering
	\caption{Object detection and tracking results of sequence \textbf{sequence3-Astra}.}
	\begin{tabular}{lrrrr}
		\toprule
		\textbf{class} & \textbf{detection} & \textbf{FP} & \textbf{FN} & \textbf{avg. error {[}m{]}}\\ 
		\toprule
		door &	18 &	1 &	12 &	0.67\\
		bench &	0  &	0 &	0 &	0\\
		trash bin &	2 &	0 &	2 &	0.47 \\
		fire exting. &	4 &	0 &	1 &	7.62 \\
		water fount. &	7 &	3 &	1 &	0.35 \\
		\bottomrule
	\end{tabular}
	\label{table-no-7th}
\end{table}

\subsubsection{Localization-based Mode} 
\begin{figure*}[!t]
	\centering    \includegraphics[width=1\linewidth]{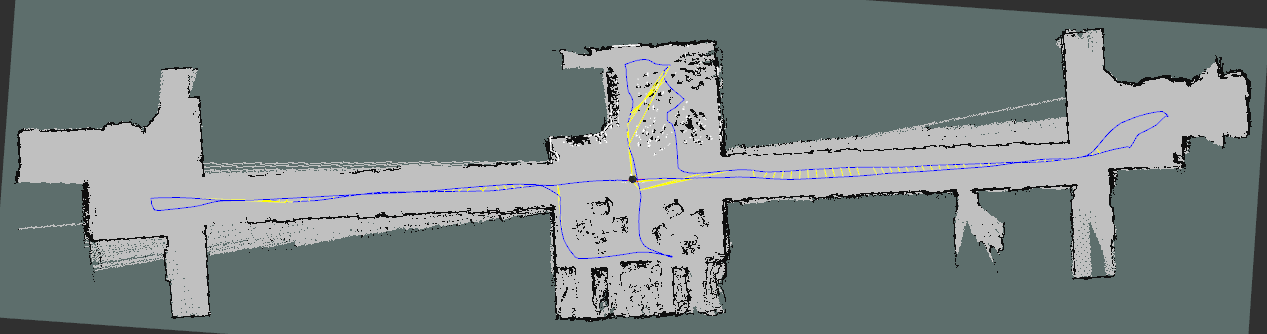}\\
	\includegraphics[width=1\linewidth]{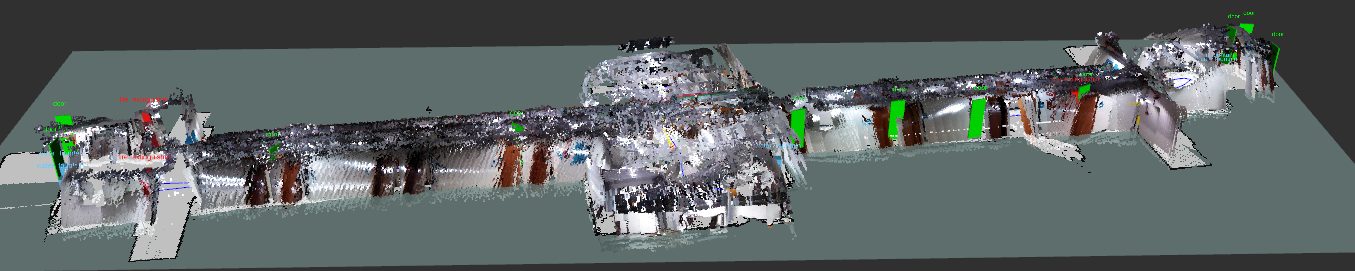}\\
	\includegraphics[width=1\linewidth]{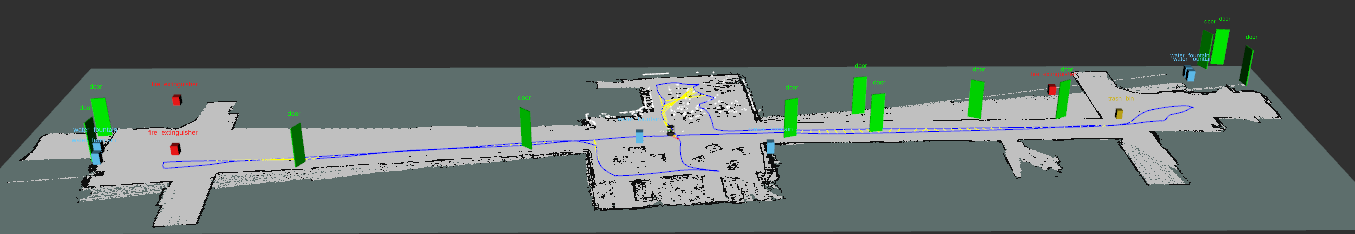}
	\caption{Visualizations of the augmented map from sequence \textbf{sequence3-Astra} with RTAB-Map. The geometric object primitives are shown in green, red and blue representing the ``door", ``water fountain" and ``fire extinguisher" respectively.}
	\label{fig:res-rtab2}
\end{figure*}

The first experiments were performed using the pure localization mode. This mode is useful for determining the accuracy of the final semantic representation since it allows the comparison of the estimated objects positions with the ground truth poses, by mitigating the undesired effects of the errors from the mapping/SLAM system back-end. The required information in this mode is a previously acquired map and a starting robot position, such as the annotated maps shown in Figures \ref{fig:ground-truth-map} and \ref{fig:ground-truth-map-7th}. One obtained simplified map view, considering solely the door objects from sequence \textbf{sequence2-Astra}, is shown in Figure \ref{fig:results1}. We then computed the number of false positives, false negatives and projected position errors as shown in Tables \ref{table-no-kine} and \ref{table-no-7th}. Note that due to hard illumination conditions, several false negatives occurred in the sequence \textbf{sequence3-Astra}, as illustrated in metrics on Table \ref{table-no-7th} and in the qualitative map visualization of Figure \ref{fig:res-rtab2}.  

We then evaluate the sensibility of the main components to sensor noise, notably affecting the RGB-D camera. We also identified some key parameters that affects directly the final obtained representation. These are the association threshold and the image detection threshold. Ideally, we would desire that the framework performance to be stable from the effects of noise and with a reasonable range of these parameters. The first performed parameter sensibility analysis is in the data association component, where we evaluated the influence of the Mahalanobis threshold to different distances as shown in Table \ref{table1}, solely for the door objects on sequence \textbf{sequence2-Astra}. We observed that small distance association values tend to cause a smaller position error, but this also favors more false positives. This effect happens since some successive object measurements were corrupted with both positioning and model extraction errors. On the other hand, large distance association values affected close-by objects to be interpreted as the same instance.

\begin{table}[t!]
	\centering
	\caption{Results varying the Mahalanobis distance threshold for the ``door'' class (\(\delta \)) for sequence \textbf{sequence2-Astra}.}
	\begin{tabular}{rrrrr}
		\toprule
		\textbf{\(\delta \) {[}m{]} } & \textbf{avg. error {[}m{]}} & \textbf{std {[}m{]}} & \textbf{FP} & \textbf{FN}\\ 
		\toprule
		
		$0.9$       & $0.46$     & $0.25$  & $27.2\%$   & $0\%$   \\
		$1.0$       & $0.70$     & $0.49$  & $18.2\%$   & $0\%$   \\
		$1.2$       & $0.54$     & $0.45$  & $11\%$    & $11\%$ \\
		$1.5$       & $0.87$     & $0.63$  & $0\%$      & $11\%$ \\
		
		\bottomrule
	\end{tabular}
	\label{table1}
\end{table}

\begin{table}[t!]
	\caption{Sensitivity experiments for different Gaussian noise levels for ``door" and ``fire extinguisher" objects using sequence \textbf{sequence1-Kinect}: (\textbf{Left}) Number of false positives (FP) and negatives (FN) of the final semantic representation by increasing noise in the RGB-D images. (\textbf{Right}) {Sampled noise trial example for the highest variance level. Due to strong appearance changes, a ``fire extinguisher" object, appearing in the left region of the image, was not detected over all frames and thus not included in the final representation.}}
	\label{table-noise}
	\begin{minipage}[b!]{.45\textwidth}	
		\begin{tabular}{l|c|c}
			\toprule
			& door & fire extinguisher\\
			$\sigma_I$ & (\textbf{FP},\textbf{FN}) & (\textbf{FP},\textbf{FN}) \\ 
			\toprule
			1 &	(0,0) & (3,2)\\		
			5 &	(1,0) & (3,2) \\
			10 &(0,2) & (2,1)\\
			20 & (1,1) & (0,4)\\
			\bottomrule
		\end{tabular}	
	\end{minipage}\hspace{0.3cm}
	\begin{minipage}[b!]{.45\textwidth}
		\includegraphics[width=1\linewidth]{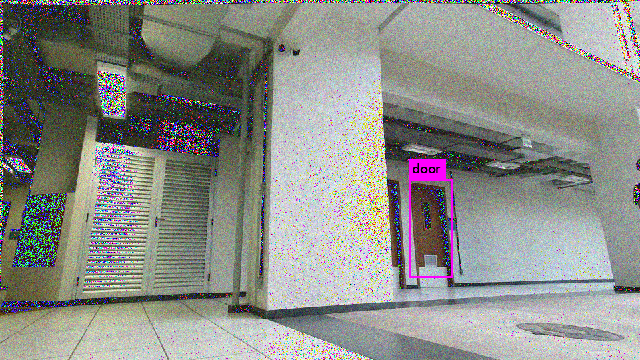}\\
	\end{minipage}
\end{table}

We then realized experiments to evaluate the system robustness to different levels of noise in the RGB-D images, with errors following the properties: 
\begin{itemize}
	\item RGB: $\tilde{\mathcal{I}}(\mathbf{p}) = \mathcal{I}(\mathbf{p}) + \mbf e_I(\mathbf{p})$ and  $\mbf e_I(\mathbf{p}) \sim \mathcal{N}\left(0,\mbf \sigma_I^2\right)\mbf{I}_{1\times3}$, for $\sigma_I \in \{1, 5, 10, 20\}$.
	\item Depth: $\tilde{\mathcal{D}}(\mathbf{p}) =  \mathcal{D}(\mathbf{p}) + e_D(\mathbf{p})$ and  $e_D(\mathbf{p}) \sim \mathcal{N}\left(0,\mbf \sigma_D^2\right)$, for $\sigma_D \in 0.1\sigma_I $.
\end{itemize}

The effects of the corrupted data in the detection, tracking and positioning components were analyzed taking ``door" and ``fire extinguisher" classes in the \textbf{sequence1-Kinect} data sequence. This choice is due the larger number of these objects that could be observed in the scene (19 and 10 respectively). These results are presented in Table \ref{table-noise}. We noted that the system was affected mainly for the larger errors with variance $\sigma_I^2 = 400$, with the increase of false positives, notably for ``fire extinguisher" class. This indicated that the detection, object filtering and tracking were capable of handling these source of errors, but the performance was deprecated for the higher noise level, as shown in the image shown on right of Table \ref{table-noise}.

\subsubsection{SLAM-based Mode}
In the SLAM mode, the extended semantic and metric maps are built concurrently, while the robot explores the scene. As described in Section \ref{sec:loc}, our formulation is adapted to use the output of some commonly employed SLAM algorithms such as Gmapping and RTAB-Map. Some obtained semantic map representations are shown in Figures \ref{fig:res-rtab1}, \ref{fig:res-rtab2} and \ref{fig:res-rtab3}. One noticed drawback of using the SLAM mode is that the formulation needs to explicitly handle the loop closing and bundle adjustment in the map generation of large spaces. While this is often done for reducing drift in trajectory errors, the tracking components should be aware of past pose adjustments to avoid misplaced objects. This undesired effect happens notably with Gmapping, which does not provide a public API of the pose graph nodes of the robot trajectory. However, this effect was greatly reduced when using RTAB-Map since we could recover the pose graph nodes directly, as illustrated in Figures \ref{fig:res-rtab1}, \ref{fig:res-rtab2} and \ref{fig:res-rtab3}.

Finally, we also considered the publicly available data sequence from RTAB-Map\footnote{demo\_mapping.bag sequence provided at \url{http://wiki.ros.org/rtabmap_ros}.} as shown in Figure \ref{fig:qual-ratb}. We note, however, that the available objects appearance were significantly altered from the trained ones and, thus, only a few door instances were observed and retained.   

\subsubsection{Discussion and Limitations}

We reduced the influence of the threshold association, presented in Table \ref{table1}, by taking into account all observed instances simultaneously with the Hungarian algorithm assignment. Still, some scenarios were affected by this parameter, as when the objects were observed while being revisited after the robot had traveled long distances in and out of the object's surrounding area in the SLAM mode. In these cases, the localization component was not capable of correcting the trajectory and mapping drifts. 
\begin{figure*}[!t]
	\centering\includegraphics[width=1\linewidth]{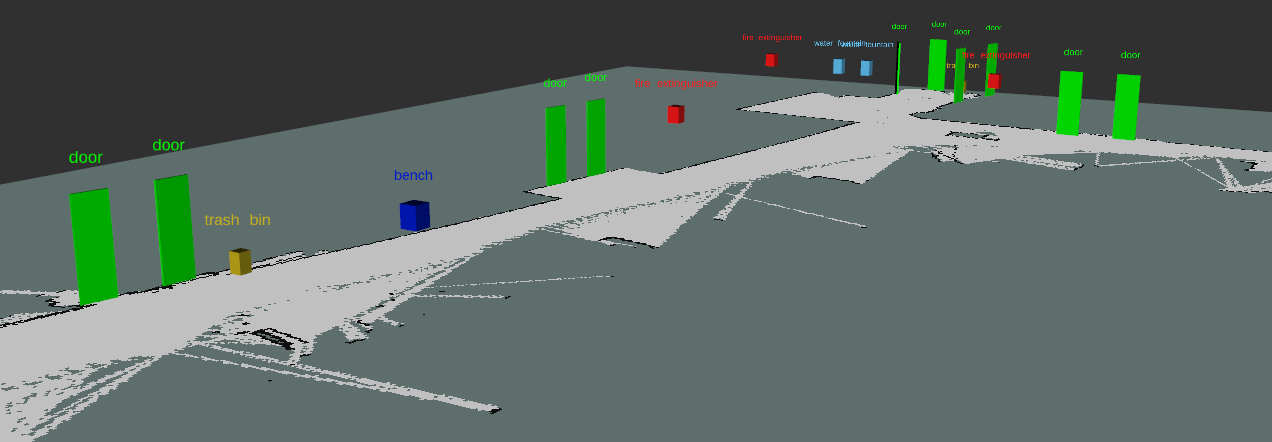}
	\caption{Augmented 2D map of objects using \textbf{sequence2-Astra} with RTAB-Map. The geometric object primitives are shown in green, red and blue representing the ``door", ``water fountain" and ``fire extinguisher" respectively. }
	\label{fig:res-rtab3}
\end{figure*}

\begin{figure*}[!t]
	\centering\includegraphics[width=0.48\linewidth]{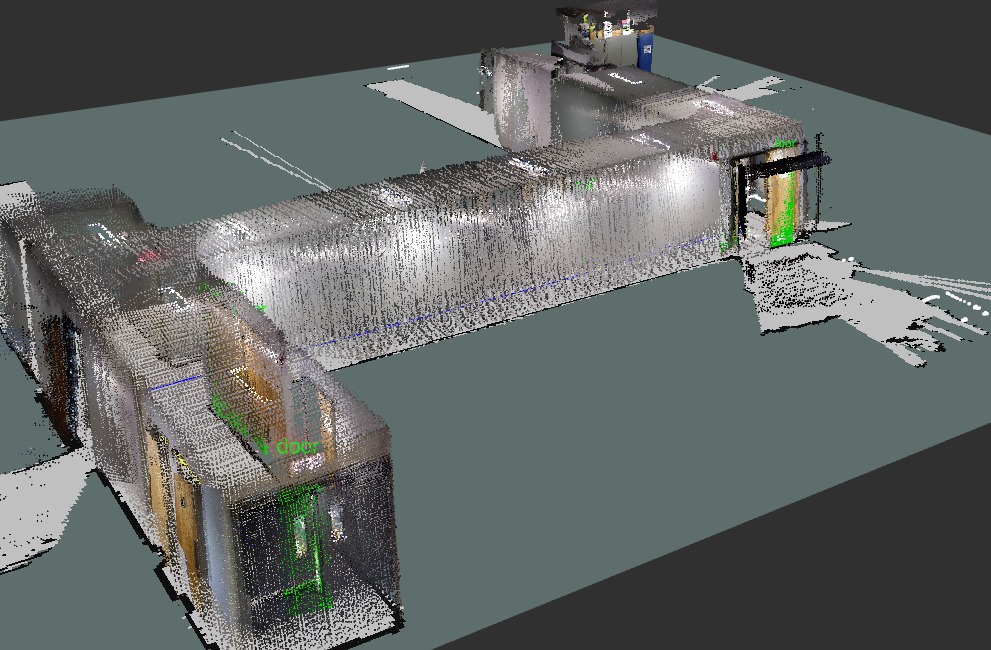}\hspace{0.2cm}\includegraphics[width=0.48\linewidth]{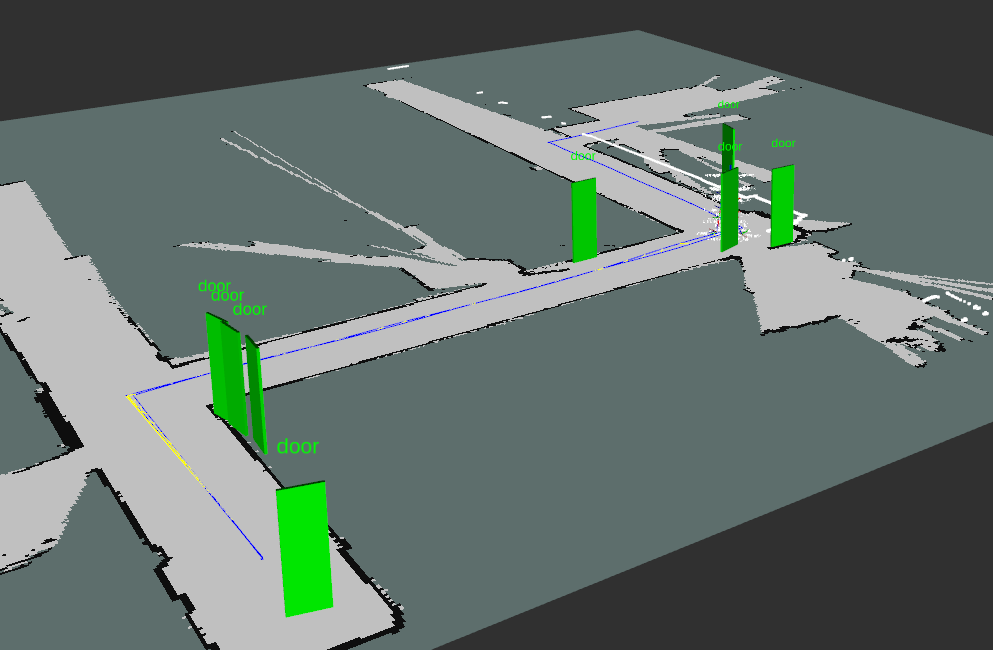}
	\caption{Augmented map results from the sequence made available in RTAB-Map. Due to appearance gap only few door objects were successfully modeled into the semantic representation.}
	\label{fig:qual-ratb}
\end{figure*}

Another parameter affecting the system performance was the latency in the object detection step, which is mainly linked with the image processing step and ROS internal inter-process communication delay. We noticed in these cases that the objects frequently were projected into bad map locations, notably when the robot performed fast rotations. We reduced the effects of this practical limitation by storing the robot pose states at the moment when the network image processing started. 

Describing each object as a two dimensional point on the map, the localization and tracking steps are greatly simplified. We might note, however, that this approach disregards object's dimensions, yielding a higher average error for larger objects, providing only a rough estimate of their position. Also, for dynamic classes (e.g., humans), tracking becomes harder, requiring more robust filtering approaches, faster processing speeds, and likely taking the object's appearance into consideration \cite{kim2018online} \cite{dimitrievski2019behavioral}.

We also found that object localization component was improved by setting a threshold on the maximum distance of projected objects from the robot. Conversely, objects that are visible from far away (greater than six to eight meters) were not taken into account, although this also incurred some loss in the object detection scores. Finally, let us conclude with an overview of the qualitative experimental results. Although the previously discussed limitations, the generated augmented semantic maps indicate desired characteristics for robot navigation and interaction tasks, as shown in the representations of the different dataset sequences in Figures \ref{fig:intro}, \ref{fig:res-rtab1}, \ref{fig:res-rtab2}, \ref{fig:res-rtab3} and \ref{fig:qual-ratb}. In the case of the SLAM mode, the object localization error was affected by the robot position error itself. When bundle adjustments of the map were performed or the robot position was corrected, previously localized instances had to be corrected as well. This requires an additional object association step for every correction, which is sometimes hard to be successfully done as shown in the augmented maps shown in Figure \ref{fig:res-rtab2}.

\section{CONCLUSIONS}\label{sec:conc}

This paper proposes a complete methodology and framework for building augmented maps with object-level information. This mapping framework is flexible and can be used with different sensor configurations, where the minimal required sensor setup consists of an RGB-D camera or a stereo camera rig. The first part of formulation leverages object detection with a shape segmentation strategy to perform instance semantic segmentation. This showed suitable for real-time operation in mobile robotic systems with limited computational resources, being an alternative to recent instance segmentation frameworks \cite{mask17,bolya-iccv2019}. The gathered information of the objects is improved overtime with a Kalman filtering tracking strategy, where the instances' associations are done using the Hungarian algorithm. The system was built on top of ROS, and it is highly modular, i.e., it can be easily modified without the need of changing other independent modules. The evaluation of the formulation was done in different indoor data sequences acquired in real conditions, containing people and objects as doors and other commonly found public space furnitures. This extended map representation can be used then with motion planning algorithms and to provide situation awareness for navigation tasks. We also provide the code and a dataset composed of three data sequences, with annotated object classes (doors, fire extinguishers, benches, water fountains) and their positioning.

A possible extension to the presented work is to consider simultaneously both color and depth information in the object instance segmentation and localization. Ideally, both the object's detection, shape and pose would be performed simultaneously, in the sense of recent formulations discussed in the works about 3D shape and pose learning from images \cite{kaolin2019arxiv,hmrKanazawa17}. Note however that our application scenarios require efficient algorithms, ideally displaying real-time performance in resource limited platforms. Another exciting direction would be to consider the semantic map in the localization while the robot navigates, as well as adopting a motion planning policy using the knowledge of the observed objects in reactive or proactive manners, seamless to how humans navigate and operates in daily-life conditions.

\section*{Acknowledgments}
The authors thank PNPD-CAPES and FAPEMIG for financial support during this research. We also gratefully acknowledge NVIDIA for the donation of the Jetson TX2 GPU used in the online experiments of this research.

\bibliographystyle{splncs}
\bibliography{main_preprint}

\begin{thebibliography}{10}

\bibitem{safety19}
Bozhinoski, D., Di~Ruscio, D., Malavolta, I., Pelliccione, P., Crnkovic, I.:
\newblock Safety for mobile robotic systems: A systematic mapping study from a
  software engineering perspective.
\newblock Journal of Systems and Software \textbf{151} (2019)

\bibitem{kit18}
Rehder, E., Wirth, F., Lauer, M., Stiller, C.:
\newblock Pedestrian prediction by planning using deep neural networks.
\newblock CoRR (2018)

\bibitem{lane18}
Carneiro, R., Nascimento, R., Guidolini, R., Cardoso, V., Oliveira{-}Santos,
  T., Badue, C., Souza, A.D.:
\newblock Mapping road lanes using laser remission and deep neural networks.
\newblock CoRR (2018)

\bibitem{pronobis12}
Pronobis, A., Jensfelt, P.:
\newblock Large-scale semantic mapping and reasoning with heterogeneous
  modalities.
\newblock In: IEEE ICRA. (2012)

\bibitem{papadakis17}
Papadakis, P., Rives, P.:
\newblock Binding human spatial interactions with mapping for enhanced mobility
  in dynamic environments.
\newblock Autonomous Robots \textbf{41}(5) (2017)

\bibitem{dbersan18}
Bersan, D., Martins, R., Campos, M., Nascimento, E.R.:
\newblock Semantic map augmentation for robot navigation: A learning approach
  based on visual and depth data.
\newblock In: IEEE LARS. (2018)

\bibitem{perez14}
P{\'e}rez-Yus, A., L{\'o}pez-Nicol{\'a}s, G., Guerrero, J.:
\newblock Detection and modelling of staircases using a wearable depth sensor.
\newblock In: ECCV. (2014)

\bibitem{leo2017computer}
Leo, M., Medioni, G., Trivedi, M., Kanade, T., Farinella, G.M.:
\newblock Computer vision for assistive technologies.
\newblock Computer Vision and Image Understanding \textbf{154} (2017)

\bibitem{wang2019self}
Wang, H., Sun, Y., Liu, M.:
\newblock Self-supervised drivable area and road anomaly segmentation using
  rgb-d data for robotic wheelchairs.
\newblock IEEE Robotics and Automation Letters \textbf{4}(4) (2019)

\bibitem{li16semi}
Li, X., Belaroussi, R.:
\newblock Semi-dense 3d semantic mapping from monocular slam.
\newblock arXiv preprint arXiv:1611.04144 (2016)

\bibitem{davidson17}
McCormac, J., Handa, A., Davison, A., Leutenegger, S.:
\newblock Semanticfusion: Dense 3d semantic mapping with convolutional neural
  networks.
\newblock In: {IEEE ICRA}. (2017)

\bibitem{hane2016}
H{\"a}ne, C., Zach, C., Cohen, A., Pollefeys, M.:
\newblock Dense semantic 3{D} reconstruction.
\newblock IEEE Trans. on Pattern Analysis and Machine Intelligence
  \textbf{39}(9) (2016)

\bibitem{wang18semantic}
Wang, C., Hou, S., Wen, C., Gong, Z., Li, Q., Sun, X., Li, J.:
\newblock Semantic line framework-based indoor building modeling using
  backpacked laser scanning point cloud.
\newblock ISPRS journal of photogrammetry and remote sensing \textbf{143}
  (2018)

\bibitem{seman2019survey}
Lateef, F., Ruichek, Y.:
\newblock Survey on semantic segmentation using deep learning techniques.
\newblock Neurocomputing (2019)

\bibitem{alexnet12}
Krizhevsky, A., Sutskever, I., Hinton, G.:
\newblock Imagenet classification with deep convolutional neural networks.
\newblock In: NIPS. (2012)

\bibitem{rcnn15}
Ren, S., He, K., Girshick, R., Sun, J.:
\newblock Faster r-cnn: Towards real-time object detection with region proposal
  networks.
\newblock In: NIPS. (2015)

\bibitem{shot15}
Liu, W., Anguelov, D., Erhan, D., Szegedy, C., Reed, S., Fu, C., Berg, A.:
\newblock Ssd: Single shot multibox detector.
\newblock In: IEEE ECCV. (2016)

\bibitem{yolo16}
Redmon, J., Divvala, S., Girshick, R., Farhadi, A.:
\newblock You only look once: Unified, real-time object detection.
\newblock In: IEEE CVPR. (2016)

\bibitem{pascal10}
Everingham, M., Van~Gool, L., Williams, C.K.I., Winn, J., Zisserman, A.:
\newblock The pascal visual object classes (voc) challenge.
\newblock International Journal of Computer Vision \textbf{88}(2) (2010)

\bibitem{ILSVRC15}
Russakovsky, O., Deng, J., Su, H., Krause, J., Satheesh, S., Ma, S., Huang, Z.,
  Karpathy, A., Khosla, A., Bernstein, M., Berg, A.C., Fei-Fei, L.:
\newblock {ImageNet Large Scale Visual Recognition Challenge}.
\newblock International Journal of Computer Vision \textbf{115}(3) (2015)

\bibitem{mask17}
He, K., Gkioxari, G., Doll{\'a}r, P., Girshick, R.:
\newblock Mask r-cnn.
\newblock In: IEEE ICCV. (2017)

\bibitem{bolya-iccv2019}
Bolya, D., Zhou, C., Xiao, F., Lee, Y.J.:
\newblock Yolact: {Real-time} instance segmentation.
\newblock In: ICCV. (2019)

\bibitem{gaidon2016virtual}
Gaidon, A., Wang, Q., Cabon, Y., Vig, E.:
\newblock Virtual worlds as proxy for multi-object tracking analysis.
\newblock In: IEEE CVPR. (2016)

\bibitem{cityscapes}
Cordts, M., Omran, M., Ramos, S., Rehfeld, T., Enzweiler, M., Benenson, R.,
  Franke, U., Roth, S., Schiele, B.:
\newblock The cityscapes dataset for semantic urban scene understanding.
\newblock In: IEEE CVPR. (2016)

\bibitem{zhan2018mix}
Zhan, X., Liu, Z., Luo, P., Tang, X., Loy, C.C.:
\newblock Mix-and-match tuning for self-supervised semantic segmentation.
\newblock In: AAAI Conference on Artificial Intelligence. (2018)

\bibitem{dai2017scannet}
Dai, A., Chang, A.X., Savva, M., Halber, M., Funkhouser, T., Nie{\ss}ner, M.:
\newblock Scannet: Richly-annotated 3d reconstructions of indoor scenes.
\newblock In: IEEE CVPR. (2017)

\bibitem{armeni17}
{Armeni}, I., {Sax}, A., {Zamir}, A.R., {Savarese}, S.:
\newblock {Joint 2D-3D-Semantic Data for Indoor Scene Understanding}.
\newblock ArXiv e-prints (2017)

\bibitem{salaris15}
Salaris, P., Vassallo, C., Sou{\`e}res, P., Laumond, J.P.:
\newblock The geometry of confocal curves for passing through a door.
\newblock IEEE Trans. on Robotics \textbf{31}(5) (2015)

\bibitem{firman16}
Firman, M.:
\newblock {RGBD Datasets: Past, Present and Future}.
\newblock In: CVPR Workshop on Large Scale 3D Data: Acquisition, Modelling and
  Analysis. (2016)

\bibitem{Nascimento:2011:IROSASP}
Nascimento, E.R., Oliveira, G., Campos, M., Vieira, A.:
\newblock {Improving Object Detection and Recognition for Semantic Mapping with
  an Extended Intensity and Shape based Descriptor}.
\newblock In: {IEEE IROS Workshop on Active Semantic Perception}. (2011)

\bibitem{moreno15}
Whelan, T., Leutenegger, S., Salas{-}Moreno, R.F., Glocker, B., Davison, A.:
\newblock Elasticfusion: Dense {SLAM} without {A} pose graph.
\newblock In: Robotics: Science and Systems. (2015)

\bibitem{engel14}
Engel, J., Sch{\"{o}}ps, T., Cremers, D.:
\newblock {LSD-SLAM:} large-scale direct monocular {SLAM}.
\newblock In: {IEEE ECCV}. (2014)

\bibitem{resnet16}
He, K., Zhang, X., Ren, S., Sun, J.:
\newblock Deep residual learning for image recognition.
\newblock In: IEEE CVPR. (2016)

\bibitem{gmapping07}
Grisetti, G., Stachniss, C., Burgard, W.:
\newblock Improved techniques for grid mapping with rao-blackwellized particle
  filters.
\newblock IEEE Trans. on Robotics \textbf{23}(1) (2007)

\bibitem{fox1999monte}
Fox, D., Burgard, W., Dellaert, F., Thrun, S.:
\newblock Monte carlo localization: Efficient position estimation for mobile
  robots.
\newblock AAAI/IAAI \textbf{1999} (1999)

\bibitem{rtab19}
Labb{\'e}, M., Michaud, F.:
\newblock Rtab-map as an open-source lidar and visual simultaneous localization
  and mapping library for large-scale and long-term online operation.
\newblock Journal of Field Robotics \textbf{36}(2) (2019)

\bibitem{orbslam15}
Mur-Artal, R., Montiel, J.M., Tardos, J.D.:
\newblock Orb-slam: a versatile and accurate monocular slam system.
\newblock IEEE Trans. on Robotics \textbf{31}(5) (2015)

\bibitem{endres14}
Endres, F., Hess, J., Sturm, J., Cremers, D., Burgard, W.:
\newblock 3-d mapping with an rgb-d camera.
\newblock IEEE Trans. on Robotics \textbf{30}(1) (2014)

\bibitem{Rusu_ICRA2011_PCL}
Rusu, R., Cousins, S.:
\newblock {3D is here: Point Cloud Library (PCL)}.
\newblock In: {IEEE ICRA}. (2011)

\bibitem{ransaccomp}
Raguram, R., Frahm, J.M., Pollefeys, M.:
\newblock A comparative analysis of ransac techniques leading to adaptive
  real-time random sample consensus.
\newblock In: European Conference on Computer Vision. (2008)

\bibitem{sadeghian2017tracking}
Sadeghian, A., Alahi, A., Savarese, S.:
\newblock Tracking the untrackable: Learning to track multiple cues with
  long-term dependencies.
\newblock In: IEEE ICCV. (2017)

\bibitem{kim2018multi}
Kim, C., Li, F., Rehg, J.M.:
\newblock Multi-object tracking with neural gating using bilinear lstm.
\newblock In: IEEE ECCV. (2018)

\bibitem{kuhn1955hungarian}
Kuhn, H.W.:
\newblock The hungarian method for the assignment problem.
\newblock Naval research logistics quarterly \textbf{2}(1-2) (1955)

\bibitem{kim2018online}
Kim, S.J., Nam, J.Y., Ko, B.C.:
\newblock Online tracker optimization for multi-pedestrian tracking using a
  moving vehicle camera.
\newblock IEEE Access \textbf{6} (2018)  48675--48687

\bibitem{dimitrievski2019behavioral}
Dimitrievski, M., Veelaert, P., Philips, W.:
\newblock Behavioral pedestrian tracking using a camera and lidar sensors on a
  moving vehicle.
\newblock Sensors \textbf{19}(2) (2019)  391

\bibitem{kaolin2019arxiv}
J., K., Smith, E., Lafleche, J.F., {Fuji Tsang}, C., Rozantsev, A., Chen, W.,
  Xiang, T., Lebaredian, R., Fidler, S.:
\newblock Kaolin: A pytorch library for accelerating 3d deep learning research.
\newblock arXiv:1911.05063 (2019)

\bibitem{hmrKanazawa17}
Kanazawa, A., Black, M.J., Jacobs, D.W., Malik, J.:
\newblock End-to-end recovery of human shape and pose.
\newblock In: Computer Vision and Pattern Regognition (CVPR). (2018)

\end{thebibliography}
\end{document}